\documentclass[11pt]{article}
\usepackage{amsmath,amssymb}
\usepackage{graphicx}
\usepackage{todonotes}
\usepackage{enumerate}
\usepackage{natbib}
\usepackage{url} \usepackage{makecell}

\usepackage[skip=0pt, font = small]{caption}
\usepackage{hyperref}
\usepackage{booktabs}
\usepackage{soul}
\usepackage{authblk}

\newcommand{\blind}{0}

\newcommand{\suppl}{0}

\newtheorem{definition}{Definition}[section]
\newtheorem{remark}{Remark}[section]

\DeclareMathOperator{\Conn}{Conn}
\DeclareMathOperator{\Comp}{Comp}
\DeclareMathOperator{\Sep}{Sep}
\DeclareMathOperator{\DCSI}{DCSI}
\DeclareMathOperator{\MST}{MST}
\DeclareMathOperator{\Dunn}{Dunn}
\DeclareMathOperator{\CH}{CH}
\DeclareMathOperator{\DB}{DB}
\DeclareMathOperator{\Sil}{Sil}
\DeclareMathOperator{\CVNN}{CVNN}
\DeclareMathOperator{\ICD}{ICD}
\DeclareMathOperator{\BCD}{BCD}
\DeclareMathOperator{\DSI}{DSI}
\DeclareMathOperator{\NOne}{N1}
\DeclareMathOperator{\NTwo}{N2}
\DeclareMathOperator{\NThree}{N3}
\DeclareMathOperator{\Density}{Density}
\DeclareMathOperator{\ClsCoef}{ClsCoef}

\DeclareMathOperator{\norm}{norm}

\begin{document}

\def\spacingset#1{\renewcommand{\baselinestretch}{#1}\small\normalsize} \spacingset{1}

\if0\blind
{
  \title{\bf DCSI - An improved measure of cluster separability based on separation and connectedness}
  \author[1,2]{Jana Gauss\footnote{
   Corresponding author email address: \texttt{jana.gauss@stat.uni-muenchen.de}}}
   \author[1,2]{Fabian Scheipl}
   \author[1,2,3]{Moritz Herrmann}
   \affil[1]{Department of Statistics, Ludwig-Maximilians-Universität München, Munich, Germany}
   \affil[2]{Munich Center for Machine Learning, Munich, Germany}
   \affil[3]{Institute for Medical Information Processing, Biometry, and Epidemiology, Faculty of Medicine, Ludwig-Maximilians-Universität München, Munich, Germany}
  \maketitle
} \fi

\if1\blind
{
  \bigskip
  \bigskip
  \bigskip
  \begin{center}
    {\LARGE\bf DCSI - An improved measure of cluster separability based on separation and connectedness}
\end{center}
  \medskip
} \fi
\bigskip
\begin{abstract}
Whether class labels in a given data set correspond to meaningful clusters is crucial for the evaluation of clustering algorithms using real-world data sets.
This property can be quantified by separability measures.
The central aspects of separability for density-based clustering are between-class separation and within-class connectedness, and neither classification-based complexity measures nor cluster validity indices (CVIs) adequately incorporate them.
A newly developed measure (density cluster separability index, DCSI) aims to quantify these two characteristics and can also be used as a CVI.
Extensive experiments on synthetic data indicate that DCSI correlates strongly with the performance of DBSCAN measured via the adjusted Rand index (ARI) but lacks robustness when it comes to multi-class data sets with overlapping classes that are ill-suited for density-based hard clustering.
Detailed evaluation on frequently used real-world data sets shows that DCSI can correctly identify touching or overlapping classes that do not correspond to meaningful density-based clusters.
\end{abstract}

\vfill

\newpage
\spacingset{1.5} 

\section{Introduction}
\noindent 
We introduce a new measure that quantifies the consistency between a given partition of a data set, e.g., as defined by a set of class labels or a cluster solution, and the underlying geometric structure of the data set. 
Our approach builds on a density-based notion of clustering \citep{hartigan1975clustering, azzalini2007clustering, chacon2015population, campello2020densityclust}, where each cluster is considered to be a connected region of higher data density that is separated from other clusters by areas of relatively lower or zero density.
Our topologically motivated understanding of (density-based) clustering considers clusters to be the connected components of a data set, which partition the data into \emph{disjoint} subsets \citep{wasserman2018}.  
More specifically, we build upon the framework by \citet{niyogi2011}: based on the manifold assumption, i.e, the assumption that the (high-dimensional) data points concentrate around a (low-dimensional) manifold embedded in the observation space, the goal of cluster analysis is to identify the connected components of this (low-dimensional) manifold.  
Consequently, in this work, we consider ``meaningful" clusters to be the \emph{connected components} of a data set which, by definition, cannot overlap or touch. 
Where it is necessary to clearly distinguish our notion of a cluster from other notions, we explicitly refer to the \emph{connected components} instead of \emph{clusters}.  

The proposed measure, the \emph{Density Cluster Separability Index} (DCSI), relies on a notion of core points similar to the popular density-based clustering algorithm DBSCAN \citep{ester} to determine relevant geometric properties of these connected components.

Why is a measure of consistency between a given partition and the underlying data structure useful and necessary? 
First of all, evaluating clustering methods frequently involves comparing the obtained clusters with the classes of a real-world data set, i.e., class labels that are supposed to represent a ``ground truth'' partition which the cluster analysis attempts to recover \citep{Elephant, Hennig}. While it is widely adopted for pragmatic reasons, this approach can be highly misleading since it is usually not known whether the partition implied by these labels results in the kind of structure that a particular clustering algorithm is designed to identify. The issue is well known in the literature -- e.g., \citet{schubert2017} suggest that the ``wrong'' data sets for evaluation might be used in many studies, since the class labels that serve as ``ground truth'' may not define a partition of the data into ``meaningful'' clusters. In a similar vein, \citet{herrmann} emphasize the necessity to differentiate between a probabilistic perspective on clustering (mixtures of distributions, ``fuzzy'' clustering) and the topological perspective we adopt here. In particular, they demonstrate that method comparisons using labeled data can be misleading if clustering methods based on these different perspectives are compared. Secondly, it is crucial that the given partition adequately reflects the desired characteristics for the specific context at hand \citep{Elephant, Hennig}. It is thus vitally important to reliably quantify the degree to which a given partition is aligned with the structure of the data, both for methodological research (e.g., in order to identify appropriate labeled data sets for benchmark studies) and in applied contexts (e.g., for evaluating specific clustering solutions or for identifying suitable clustering algorithms for a given data set).
Note that we consider separability as first and foremost a property of a given data set, and not a property of the underlying data generating process.

The new measure, DCSI, is intended to address limitations inherent in many of the existing \emph{data complexity measures} and \emph{Cluster Validity Indices} (CVIs). In addition to between-class separation, which is defined as the minimal distance occurring between core points of different classes, it also incorporates a measure of within-class connectedness (i.e., how closely the data points of a given class are connected) as a central characteristic. 
One important consequence of this approach is that the DCSI has no implicit preferences for specific cluster shapes. This is an advantage over many existing CVIs like Dunn \citep{Dunn}, CH \citep{CH} or the Silhouette index \citep{Sil}, as these measures tend to favor clusters of spherical shape by emphasizing cluster compactness (i.e., the dispersion of the data).

The remainder of the paper is structured as follows: 
We provide some intuition and background on the notion of separability and CVIs in Section \ref{sec:RelW}.
Section \ref{sec:DCSI} then defines the DCSI. 
In Section \ref{sec:ExistSep}, we compare DCSI to existing separability measures, indicating that DCSI is able to overcome their difficulties in quantifying the separability of density-based clusters. 
The results of extensive experiments on synthetic and real-world data are reported in Section \ref{sec:exp}. 
Finally, we discuss the results and presents our conclusions in Section \ref{sec:disc}.
Additional information is provided in the Appendix.
\begin{remark}
During the review process, it was pointed out to us that DCSI is very similar to two existing CVIs for density-based clusters: \emph{density-based clustering validation index} (DBCV) by \citet{DBCV} and \emph{density-core-based clustering validation index} (DCVI) by \citet{DCVI}.
At the time of writing, we were not aware of the existence of these papers.
DCSI is essentially the same as DCVI computed on core points.
DCVI can be seen as a special version of DBCV, which relies on the same ideas but uses a somewhat sophisticated density-based distance instead of euclidean distances.
Additional information on the validation of arbitrarily shaped clusters can be found in the recent survey paper by \citet{Schlake}.
For the synthetic experiments, we included two versions of DBCV: the original version by \citet{DBCV} as well as DBCV evaluated on core points only, similar to DCSI.
\end{remark}
 \section{Background}\label{sec:RelW}

\subsection{Separability}\label{sec:sepClust}

The term \textit{separability} is mainly used in the context of classification and is based on the idea that the performance of a classifier depends on two aspects: the capacity of the classifier and the separability of the data set \citep{GuanSep}.
\citet{Fer2018} describe separability as an (intrinsic) characteristic of a labeled data set that quantifies how much the classes defined by the labels overlap.
A closely related concept is \textit{complexity}, i.e., the difficulty of the induced classification problem \citep{Ho2002}.
Complexity measures map a labeled data set to a real number that quantifies this characteristics.
An overview and categorization of these measures can be
found in \citet{Lorena}.

\begin{figure}[h]
\begin{center}
\includegraphics[width=4.8in]{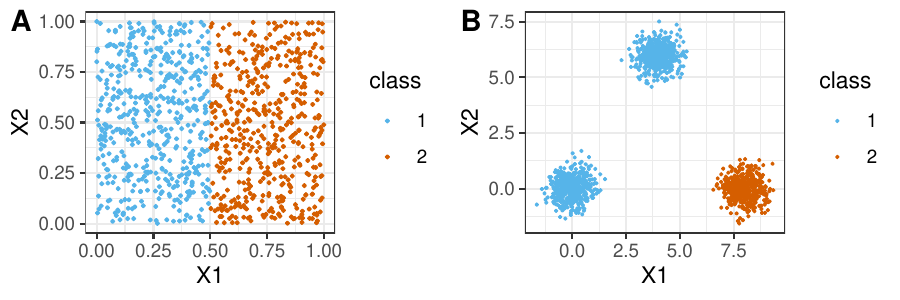}
\end{center}
\caption{Separability from a classification- vs clustering-based view\label{fig:ClusterVsClass}}
\end{figure}

Consider Figure \ref{fig:ClusterVsClass}, which illustrates two scenarios with high separability (i.e., low complexity) when observed from a classification standpoint.
In both scenarios, the two classes can easily be separated by a single linear decision boundary. Yet, the classes do not correspond to meaningful clusters in the topological sense adopted here: clusters are the connected components of a data set and therefore do not overlop or touch. In \textbf{A}, there is only one connected component of uniformly distributed data. In \textbf{B}, while the two classes are well separated, the data in class 1 (blue) are spread over two different connected components, not just one. 

A measure of separability of a data set in terms of cluster analysis needs to take both of these aspects into account. That is, it needs to take into account not only \emph{between-cluster} separation, like a complexity measure for classification does, but also \emph{within-cluster} connectedness.
It also means that separability in a clustering context requires a stricter notion of separation: In order to form meaningful clusters in the strict topological sense of connected components, the domains of different classes must not touch. 
The two examples in Figure \ref{fig:ClusterVsClass} show that a high degree of separability with regard to a classification algorithm does not guarantee that the classes as predefined by a set of labels are consistent with the connected components of a data set. 
In other words, the classes do not correspond to topologically meaningful clusters, which require both between-cluster separation and within-cluster connectedness. The DCSI, the measure we introduce in Section \ref{sec:DCSI}, aims to take into account both of these aspects. This is a crucial difference to many existing separability measures, whose suitability as separability measures is analyzed in Section \ref{sec:ExistSep}.

\subsection{CVIs}
In order to choose between competing clustering solutions or to tune the hyperparameters of a cluster analysis algorithm, it is necessary to evaluate the quality of a partition of a data set \citep{Hu, GuanCVI, LiuEnh}. 
External validation uses (true) class labels and quantifies the quality of a clustering by its agreement with such labels. 
For real-world clustering problems, class labels are not available, so internal validation is usually the only option \citep{Hu}. 
An \textit{internal cluster validity index} (CVI) uses only the predicted labels and the data  \citep{GuanCVI}. 
The term \textit{clustering quality measure} (CQM) \citep{Acker2008} is used synonymously.
A CVI is a function that maps a clustering and the data to a real number indicating how ``strong" or ``conclusive" the clustering is \citep{Acker2008}. 
A classification of some cluster validity indices can be found in \citet{Hu}. 

\citet{GuanCVI} propose to use their separability measure DSI as a CVI.
This makes sense as the aim of clustering can be described as ``finding a partition with high separability'' and the role of CVIs can be interpreted as quantifying the degree of separability of a given partition.
Conversely, it seems reasonable to use CVIs as separability measures.
The advantages and disadvantages of some popular CVIs when used as separability measures are investigated in Section \ref{sec:ExistSep}.

 \section{DCSI - a measure of separability based on connectedness and separation}
\label{sec:DCSI}

In this section, we introduce the \textit{Density Cluster Separability Index}, which aims to measure the degree to which a given partition of a data set aligns with density-based clusters, i.e., the connected components of the data.
This index is designed to quantify both \textit{separation} (how well are the classes separated from each other) and \textit{connectedness} (how well are the points within one class connected).
Similar to many CVIs being defined as ratios between measures of separation and measures of compactness \citep{LiuEnh}, this newly proposed index is based on a ratio of a measure of separation and a measure of connectedness. 
Section \ref{sec:DefDCSI} outlines their development, and Section \ref{sec:paramDCSI} discusses their computational complexity and the choice of hyperparameters.

DCSI relies on a notion of core points similar to the popular density-based clustering algorithm DBSCAN \citep[\emph{Density Based Spatial Clustering of Applications with Noise},][]{ester}.
The core idea of DBSCAN is that clusters are constituted by areas of high data density. These high density areas are separated by areas of noise, whose density is lower than the density in any of the clusters \citep{ester, schubert2017}.
DBSCAN requires two parameters, $\textit{MinPts}\in \mathbb{N}$ and $\varepsilon>0$.
Points whose $\varepsilon$-neighborhood contains a minimum number of points, \textit{MinPts}, are called \textit{core points}.
All points within the $\varepsilon$-neighborhood of a core point are assigned to the same cluster.
If any of these points is a core point, its neighbors are also included \citep{ester, schubert2017}.

\subsection{Definition of DCSI}
\label{sec:DefDCSI}
We first present a two-class version (with classes $C_1, C_2$) of DCSI.
Define hyperparameters $\textit{MinPts} \in \mathbb{N}$ and $\varepsilon_i > 0$ for each class $C_i$ and a distance metric $d(x, x')$.
Similar to DBSCAN, DCSI sets up a notion of core points: a point $x \in C_i$ is a core point if at least \textit{MinPts} observations from $C_i$ lie in its $\varepsilon_i$-neighborhood:
\begin{definition}[Core points DCSI]
The set of core points $\mathcal{C}_i$ of a class $C_i$ for given $\varepsilon_i$ and \textit{MinPts} is defined as
$ \mathcal{C}_i = \{x \in C_i: |\mathcal{N}_{\varepsilon_i}(x)| \geq \textit{MinPts} \}$, where $ \mathcal{N}_{\varepsilon_i}(x)= \{x' \in C_i\setminus \{ x \}: d(x, x')\leq \varepsilon_i \}$ for $x \in C_i$.
\end{definition}
Note that core points are calculated separately for each class: $\varepsilon_i$ is specific to each $C_i$ (different from DBSCAN) and the $\varepsilon_i$-neighborhood $\mathcal{N}_{\varepsilon_i}(x)$ of a point $x \in C_i$ contains only observations from $C_i$.  
A possible choice of $\varepsilon_i$ is described later. 
\textit{MinPts} is set up as a global parameter, but it could also be chosen for each class.

A DCSI that is based on all points is a special case of this definition: if the $\varepsilon_i$ are sufficiently large or if $\textit{MinPts}$ is 0, every point becomes a core point.

\paragraph{Separation:}
Relying on a limited set of representative data points such as class centers to quantify separation often fails; e.g., two nested spheres could have the same center.
Metrics based on mean distances between classes or nearest neighbors (e.g., the complexity measures N1, N2 and N3, see the appendix for their definitions) can also display undesirable behavior, for example in the following setting:
Imagine a linearly separable one-dimensional data set with a null margin (e.g., class 1: $x > 0$, class 2: $x \leq 0$) drawn uniformly from an interval $[-a,a]$.
As the classes touch, they are not separable from a clustering standpoint.
However, such measures would indicate higher separability as the interval expands, due to an increase in the mean distance to the nearest neighbor from a different class or a decrease in the proportion of points whose nearest neighbor belongs to a different class.
From a clustering perspective, however, the (lack of) separability remains unchanged.
Taking the minimal distance between classes into account could avoid this issue, but such an approach is too sensitive to outliers.
Therefore, a different notion of ``minimal distance'' between classes is required.
Selecting a low quantile of interclass pairwise distances is robust to outliers but has the same weakness as the measures mentioned earlier: increasing the interval width leads to an undesirable increase in apparent separability.
\begin{definition}[Separation DCSI]
\[
\Sep_{\DCSI} = \min_{x \in \mathcal{C}_1, x' \in \mathcal{C}_2} d(x,x').
\]
\end{definition}
Our proposal to attain a robust minimum distance is based on using only the core points $\mathcal{C}_i$, thereby defining the separation between the classes $C_1$ and $C_2$ as the minimal distance among core points $x \in \mathcal{C}_1, x' \in \mathcal{C}_2$. This measure of separation is fairly robust to outliers by construction and does not change when observations that are irrelevant for separability are added to the data.

\paragraph{Connectedness:}
Connectedness should be distinguished from compactness, which is typically measured based on maximum or mean distances within clusters and therefore favors classes of more spherical shape.
In order to obtain a measure that reflects the degree of within-class connectedness even if the data forms non-compact shapes like circles, a different notion of ``maximum distance'' within a class is needed.

Our suggested solution is to use the biggest distance in a minimum spanning tree (MST) connecting only the core points of a given class:

\begin{definition}[Connectedness DCSI]
\[
\Conn_{\DCSI}(C_i) = \max_{(x, x') \in V_i} d(x, x'),
\]
where $V_i$ is the set of vertices of $ \MST(\mathcal{C}_i)$, a minimum spanning tree built only from the core points $\mathcal{C}_i$ of class $C_i$.
\end{definition}
If the MST were to be constructed on the fully connected (i.e., complete) graph of the respective class members, the maximal edge weight of the MST of each class would be very sensitive to outliers and, as such, a poor indicator of intra-class connectivity.
Some high quantile of the edge weights (for instance, the 95\%-quantile) could be used instead of the maximum to get around this, but this would also fail to reliably measure connectedness -- for example in the case of a class consisting of two components (as depicted in Figure \ref{fig:ClusterVsClass} \textbf{B}) in which a single exceedingly large edge weight connects these two components.
As before, we solve these issues by focusing on the core points of each class: the relevant MST is based on the complete graph of these core points only and its largest edge weight is adopted as the metric for connectedness within a class.

This is identical to the maximum path-based distances defined in \cite{Hu} and \cite{PBC}, however, \cite{Hu} use the average path-based distance for their CVI. 
In order to obtain a value for the entire (two-class) data set, we take the maximum of $\Conn_{\DCSI}(C_1)$ and $\Conn_{\DCSI}(C_2)$:
\[
\Conn_{\DCSI} = \max\{ \Conn_{\DCSI}(C_1), \Conn_{\DCSI}(C_2)\}.
\]
This maximum is easier to interpret than the average: it is the largest distance occurring in both MSTs.

\paragraph{DCSI:}
Higher values of $\Sep_{\DCSI}$ and smaller values of $\Conn_{\DCSI}$ indicate better separability\footnote{It might be confusing that we define connectedness such that smaller values indicate better connectedness, so defining connectedness as the inverse of our proposed metric would be more intuitive. However, we decided to emphasize the similarity to some existing CVIs (like CH and Dunn), which are ratios of measures of separation and compactness \citep{LiuEnh}.}.
Similar to many CVIs in Section \ref{sec:ExistSep}, we use the quotient of separation and connectedness as our measure of separability and rescale it to $[0,1[$:
\begin{definition}[DCSI, pairwise]
\[
\DCSI  = \frac{q}{1 + q}, \text{ where } q =  \frac{\Sep_{\DCSI}}{\Conn_{\DCSI}} .
\]
\end{definition}
$\DCSI \to 0$ if $\Sep_{\DCSI} \to 0$ or $\Conn_{\DCSI} \to \infty$ and $\DCSI \to 1$ for $\Sep_{\DCSI} \gg \Conn_{\DCSI}$, i.e., if the minimum distance between core points of different classes is much higher than the maximum path-based distance between core points that belong to the same class.
A DCSI of $0.5$ indicates that $\Sep_{\DCSI} = \Conn_{\DCSI}$. 

The DCSI of a data set with more than two classes could be defined as a summary of the pairwise DCSIs, e.g., the mean, median or minimum pairwise DCSI. Another possibility is to define separation and connectedness of the entire data set as summaries of separation and connectedness of its classes.
However, this ignores the interplay between separation and connectedness of a pair of classes and can therefore lead to an overly sensitive measure.
Since it is reasonable to take all values of pairwise DCSI into account, we suggest using the mean pairwise DCSI as a measure of separability of the entire data set:

\begin{definition}[DCSI, multi-class]
Let $X$ be a data set with classes $C_1, \ldots, C_K$ and let $\DCSI(C_i, C_j)$ be the pairwise DCSI of classes $C_i$ and $C_j$.
The DCSI of the data set is given by
\[
\DCSI (X) = \frac{2}{K \cdot (K-1)} \sum_{i = 1}^{K-1}  \sum_{j = i+1}^K \DCSI(C_i, C_j) .
\]
\end{definition}

See Appendix \ref{sec:AppMultiClass} for a discussion and evaluation of other methods to define a multi-class version of DCSI.
In practice, the specifics of each application will determine which properties of a multi-class DCSI are relevant or desirable, and our experimental results in Section \ref{sec:RW} show that it often makes sense to consider not just the aggregate DCSI of the entire multi-class data set but to also investigate pairwise separabilities.

\subsection{Computational Aspects and Choice of Parameters}
\label{sec:paramDCSI}
\paragraph{Runtime complexity of DCSI:}
The time complexity is dominated by the computation of core points and the calculation of the distance matrix, which are both $O(n^2)$ in the worst case, where $n$ is the size of the data set.
The distance matrix is only needed for all core points (so in the worst case for each point): for the computation of $\Sep_{\DCSI}$ (distances between all core points of different classes) and for the construction of the MSTs (distances between all core points within one class).
The computation of core points requires $O(n^2)$, as a neighborhood query ($O(n)$) is performed for each of the $n$ data points (this is the same as the worst case runtime for DBSCAN, see \citet{schubert2017} for details).
Computing a MST for a given distance matrix of $n_i$ points (size of the $i$-th class) requires $O(n_i \log n_i)$ using Kruskal's algorithm \citep{kruskal, algo}\footnote{There is a faster algorithm achieving almost linear runtime \citep{Chaz}}, so in the worst case, the computation of all MSTs requires $O(n \log n)$\footnote{Assume there are $K$ classes. It holds $\sum_{i = 1}^K n_i \log n_i \leq \sum_{i = 1}^K n_i \log n = n \log n$, since $\sum_{i = 1}^K n_i = n$.}.

This time complexity is similar to many other separability measures, as many of them rely on the distance matrix. 
The complexity measures used in Section \ref{sec:ExistSep} are all $O(n^2)$ \citep{Lorena}, whereas the CVIs require $O(n^2)$ or $O(n)$ \citep{CVICompl}.
We assumed the number $d$ of features to be fixed. 
Taking the dimensionality of the data into account leads to a complexity of $O(d n^2)$ for the distance matrix, so the time complexity of DCSI and most other measures is $O(d n^2)$ (and $O(d n )$ for some CVIs).

\paragraph{Choice of parameters:}
A threshold parameter $\varepsilon_i > 0$ for each class $C_i$ and $\textit{MinPts} \in \mathbb{N}$ needs to be set in order to define core points. 
There is no ``true'' or ``best'' choice of these parameters, since suitable and meaningful values always depend on the specific application and would -- ideally -- be chosen based on domain knowledge, similar to DBSCAN \citep{schubert2017}.
This section aims to give insight into the effect of \textit{MinPts} and $\varepsilon_i$ and provides some guidelines for their choice. 

Recall that a point $x \in C_i$ is a core point, if it has at least \textit{MinPts} observations from $C_i$ in its $\varepsilon_i$-neighborhood.
One obtains fewer core points by increasing \textit{MinPts} for a fixed $\varepsilon_i$ or by decreasing $\varepsilon_i$ for a fixed \textit{MinPts}.
The effect of fewer core points on $\Sep_{\DCSI}$ is clear: it increases because the minimum distance between core points of different classes increases.
The effect on $\Conn_{\DCSI}$ is more complex: both an increase or a decrease in connectedness are possible. 
An increase in connectedness (i.e., a lower(!) value of $\Conn_{\DCSI}$) is observed, if a group of ``outliers'' loses their status as core points, thereby decreasing the maximum edge weight in the MST.
On the other hand, a decrease in connectedness (i.e., higher $\Conn_{\DCSI}$) is also possible, if the smaller number of core points leads to separation within a class, which increases the maximum edge weight in the MST. This effect is shown in more detail later. 

$\textit{MinPts}+1$ can be interpreted as the minimal cluster size, so that an isolated set of close points with at least this many members is not discarded as ``outliers'' or noise points and therefore affects the separability.
In this paper - unless otherwise stated - $\textit{MinPts}=5$ is always used, similar to DBSCAN \citep{hahsler2019}.
However, for very noisy or large data sets, it might make sense to choose a higher value to enhance the robustness of DCSI, which is investigated in more detail in Appendix \ref{App:RW:MinP}.
If the class sizes differ greatly, one could also consider choosing \textit{MinPts} separately for each class.

The choice of $\varepsilon_i$ is more challenging, since the range of meaningful values depends on the distances within classes.
As the densities in different classes can vary widely, a single global $\varepsilon$ can lead to the effect that some classes with lower density (i.e., higher distances) have no core points at all, so $\varepsilon_i$ is set for each class separately. If no domain knowledge is available, we suggest choosing $\varepsilon_i$ based on the distribution of the observed distances. 

For the remainder of this paper, we chose to set $\varepsilon_i$ to the median distance between points $x \in C_i$ and their $(2\cdot\textit{MinPts})$-th nearest neighbor in $C_i$.
This heuristics works well empirically (see Section \ref{sec:exp}) and seems to offer a good compromise for obtaining a reasonable amount of core points.
\begin{definition}[Proposed choice of $\varepsilon_i$]
\[
\varepsilon_i = \operatorname{median}_{x_j \in C_i} d(x_j, x_{(j, \textit{MinPts}\cdot2)}),
\]
where $ x_{(j, k)} $ denotes the $k$-th nearest neighbor of $x_j$ in $C_i$.
\end{definition}
In order to calculate the connectedness within a class, at least two core points are needed, and the proposed choice of $\varepsilon_i$ ensures that this is the case for each class\footnote{Assume that each class has at least $(2\cdot\textit{MinPts}) + 1$ data points (otherwise, the $(2\cdot\textit{MinPts})$-th nearest neighbor is not defined). Since the proposed choice of $\varepsilon_i$ is the median of $(2\cdot\textit{MinPts})$-th nearest neighbor distances, it holds that $d(x_j, x_{(j, 2 \cdot \textit{MinPts})}) \leq \varepsilon_i$ for at least $50\%$ of the data points in $C_{i}$, which also implies $d(x_j, x_{(j, \textit{MinPts})}) \leq \varepsilon_i$ for at least $50\%$ of the data points $C_{i}$, so at least $50\%$ of the points in $C_i$ are core points.}. 
Furthermore, unlike the mean, the median is robust to outliers.

\begin{figure}
\begin{center}
\includegraphics[width = \textwidth, height = 7cm]{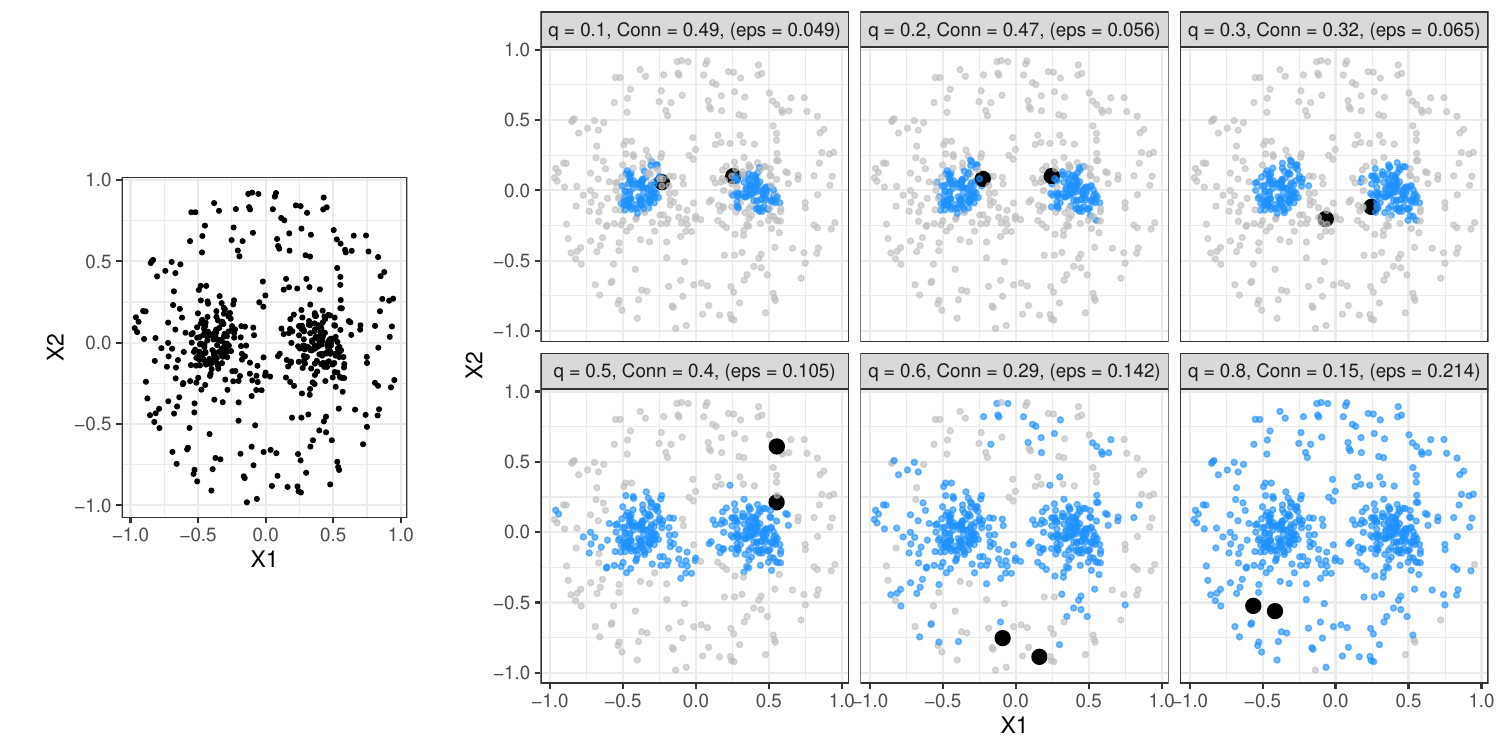}
\end{center}
\caption{Data of a class with two modes, $n = 500$ (left) and core points and connectedness for different choices of $\varepsilon$ (right). $\varepsilon$ is the $q$-quantile of the distances to the 10th nearest neighbor for $q \in \{0.1, 0.2, 0.3, 0.5, 0.6, 0.8\}$. The obtained core points (with $\textit{MinPts} = 5$) are shown in blue and the two core points that determine the connectedness are shown black. This example emphasizes that there are no ``true'' values of $\Sep$, $\Conn$ and $\DCSI$ and therefore no globally applicable ``right'' or ``optimal'' choice of the parameters.\label{fig:Par1}}
\end{figure}

In Figure \ref{fig:Par1}, alternative choices for $\varepsilon_i$ and their effects on connectedness in an exemplary data set is shown.
The data consists of one class, since this example focuses on connectedness. 
The data is sampled from a disk and two normal distributions.
Since there are two modes separated by an area of lower density, one could argue that these data are not connected. 
Alternatively, these data could be seen as one connected component, since the disk connects the two modes.
This data set shows that separability and therefore meaningful values of $\varepsilon_i$ and \textit{MinPts} will often depend on the specific application.
The second plot in Figure \ref{fig:Par1} shows the obtained core points (in blue) and the computed values of connectedness for different values of $\varepsilon_i$. 
The two core points that determine the connectedness (i.e., which are connected by the longest edge in the MST) are shown in black.
The values of $\varepsilon_i$ are chosen as the $q$-quantile of the distances to the 10th nearest neighbor (i.e., $(2\cdot\textit{MinPts})$) in the class, for $q \in \{0.1, 0.2, 0.3, 0.5, 0.6, 0.8\}$.
(A plausible alternative strategy, leading to a similar range of $\varepsilon_i$ values, would have been to set $\varepsilon_i$ to the median distance to the $k$-th nearest neighbor for different values of $k$.)

\begin{figure}
\begin{center}
\includegraphics[width = \textwidth, height = 4cm]{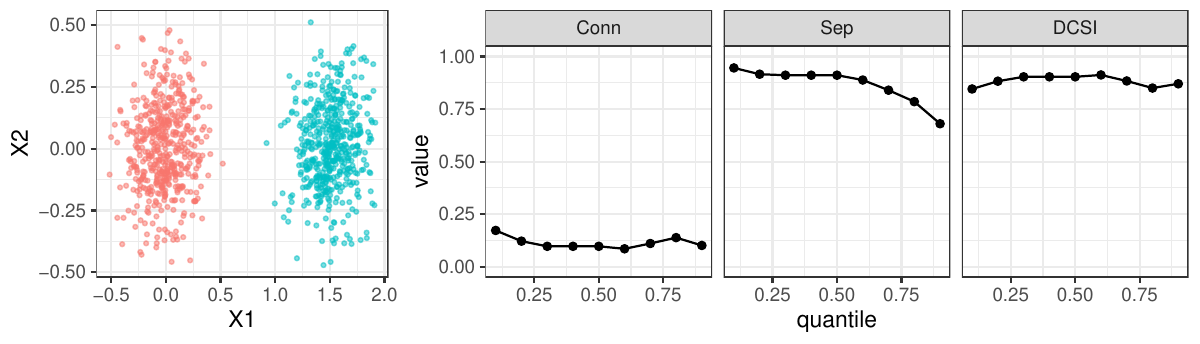}
\end{center}
\caption{Well separated two-class data set, $n_1 = n_2 = 500$ (left) and obtained values of connectedness, separation and DCSI for different $\varepsilon_i$ (right). $\varepsilon_i$ is the $q$-quantile of the distances to the 10th nearest neighbor for $q = 0.1, 0.2, \ldots, 0.9$. For these clearly separated clusters, the dependence of the measures on the specific hyperparameter values is very small. \label{fig:Par2}}
\end{figure}

One can observe the effects explained earlier: As $q$ and therefore $\varepsilon_i$ increases, the number of core points decreases, which can both lead to higher or lower connectedness:
The connectedness is worse (i.e., higher) for $q = 0.5$ compared to $q=0.3$, since a core point that is separated from the two modes emerges. 
However, the connectedness is better for $q=0.8$ compared to $q = 0.3$, as the separation between the two modes has vanished.

The example in Figure \ref{fig:Par2} and its comparison with Fig.~\ref{fig:Par1} suggest that strong dependence on the parameters will mainly occur in data with ambiguous cluster structure and that the effect of the parameters on DCSI is much smaller if the data fits the ``topological perspective'', i.e., if it concentrates around a manifold that consists of clearly distinct connected components.

In practice, it might make sense to try different $\varepsilon_i$ and investigate the variability of the resulting DCSI values. 
The examples above show that, if DCSI is strongly affected by $\varepsilon_i$, this indicates that the separability of the given classes is ambiguous and depends on the context.
 \section{Comparison to existing measures of separability \& toy example}

\label{sec:ExistSep}

This section compares DCSI to some existing measures that can be used to assess separability.
Definitions of all these measures are provided in the appendix.
Some widely used CVIs are included here, as well as a selection of complexity measures.
Some complexity measures in \citet{Lorena} are not suitable for measuring separability from a clustering-based view, e.g., linearity or class imbalance measures.
The complexity measures presented here all belong to the categories \textit{neighborhood measures} and \textit{network measures} (see the appendix for more details).
The third category in Table \ref{tab:Sep}, \textit{distributional}, is a different approach to quantifying separability: 
One can measure to what extent points from different classes mix with each other, i.e., one quantifies the dissimilarity of distributions.

Recall from Section \ref{sec:sepClust} that a separability measure for density-based clusters has to measure connectedness (Figure \ref{fig:ClusterVsClass}, \textbf{B}). Additionally,  it has to measure separation from a (density) clustering-based view, i.e., the domains of different classes must not touch or overlap in order to form meaningful density-based clusters (Figure \ref{fig:ClusterVsClass}, \textbf{A}).
The existing measures are evaluated with regard to these two aspects (column ``clustering-based'', Table \ref{tab:Sep}).
Furthermore, a separability measure should not favor convex classes but allow for arbitrary shapes (column ``arbitrary shape'').

\begin{table}

\caption{Overview of existing separability measures. As it is desirable that all measures take on values in $[0,1]$ with 1 indicating highest separability, some measures are slightly modified which is indicated by an asterisk. CVI = cluster validity index, Distr.=Distributional, Gr./Nb.=Graph-/Neighborhood-based. 
``partially'' means that a measure fulfills parts of the requirements of ``Clustering-based'' and ``Arbitrary shape'', but ignores certain aspects. See the text for explanations.
More details on the characteristics of the measures can be found in \citet{MA}.
\label{tab:Sep}}
\centering
\begin{tabular}[h]{|l|p{40mm}|l|p{20mm}|p{20mm}|l|}
\toprule
Measure & Reference & Category & Clustering-based & Arbitrary shape  \\
\midrule
$\text{Dunn}^*$ & \citet{Dunn}& CVI & yes & no  \\
$\text{CH}^*$& \citet{CH}& CVI & partially & no   \\
$\text{DB}^*$ & \citet{DB}& CVI & partially & no    \\
$\text{Silhouette}^*$ & \citet{Sil}& CVI & partially & no    \\
$\text{CVNN}^*$ & \citet{LiuEnh}& CVI & yes & partially  \\
DSI& \citet{GuanNN}& Distr. &no & yes   \\
N1& \citet{Lorena}& Gr./Nb.& no & yes   \\
N2 & \citet{Lorena}& Gr./Nb.& no & yes  \\
N3& \citet{Lorena}& Gr./Nb.& no & yes  \\
LSC& \citet{Lorena}& Gr./Nb.& partially& no   \\
Density& \citet{Lorena}& Gr./Nb.& no & partially  \\
ClsCoef& \citet{Lorena}& Gr./Nb.& no & partially   \\
$\text{DBCV}^*$& \citet{DBCV}& MST & yes & yes   \\
\bottomrule
\end{tabular}
\end{table}

Most of the existing CVIs measure compactness of classes instead of connectedness, by taking the maximum distance (Dunn), the variance (CH) or the average distance (Silhouette) within classes into account.
They therefore favor classes of spherical shape.
Furthermore, some measures (CH, DB) take distances of class centers into account in order to measure separation, which is unsuitable for arbitrarily shaped classes, e.g., concentric spheres.

As they measure not only separation but also compactness, the CVIs represent a clustering-based view of separability.
However, most of them (except Dunn) are not able to detect touching classes as in Figure \ref{fig:ClusterVsClass} \textbf{A}, so CH, DB and Silhouette are only partially clustering-based. 

CVNN aims to overcome some disadvantages of existing CVIs \citep{LiuEnh}.
Instead of cluster centers, it uses nearest neighbors to quantify separation, which makes it more suitable for arbitrarily shaped classes than the classic CVIs.
However, its notion of compactness (average pairwise intra-class distance) still favors classes of spherical shape.

DSI and the complexity measures N1, N2 and N3 are suited for arbitrarily shaped classes but they only measure separation and do not take connectedness into account, thereby representing a classification-based view.
Furthermore, if additional points distant from the border were added in Figure \ref{fig:ClusterVsClass} \textbf{A}, these measures would indicate a higher separability even though the data would not be easier to separate (from a clustering-based view) than before.

LSC favors spherical classes and measures the compactness of the classes to some extent, so it is neither clearly classification- nor clustering-based.
The network measures Density and ClsCoef slightly favor convex classes and measure neither connectedness nor compactness.\\

\begin{figure}
\begin{center}
\includegraphics[width = 4.8in]{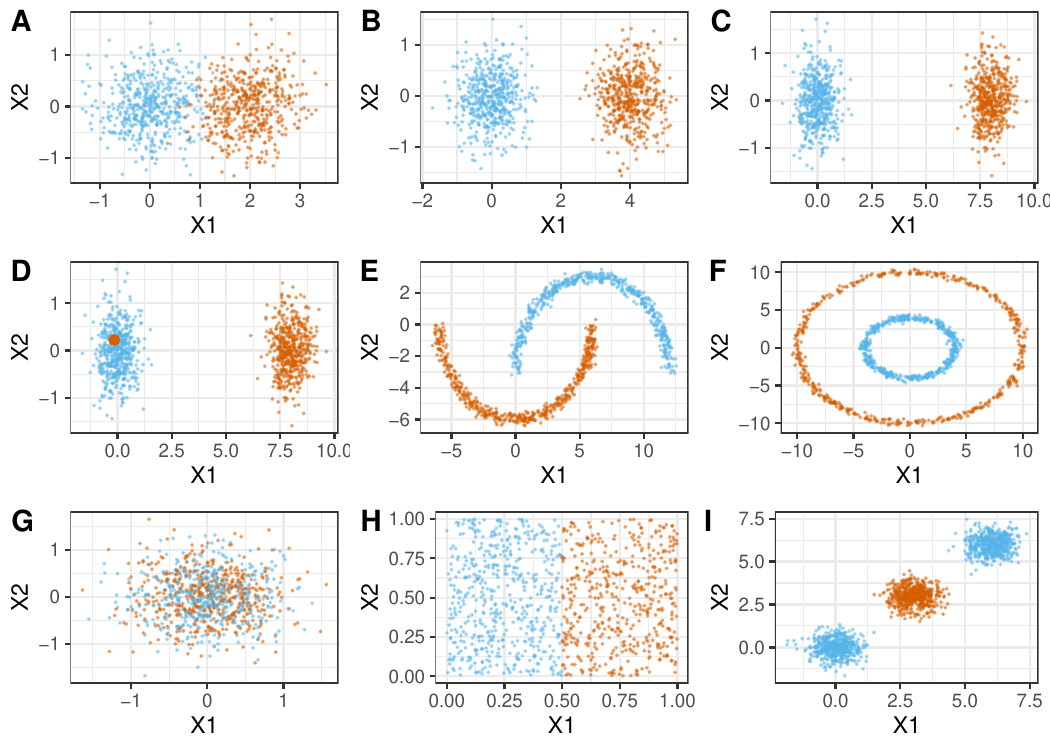}
\end{center}
\caption{Exemplary data sets to evaluate separability measures \label{fig:SepExmp}}
\end{figure}

\begin{table}
\caption{Existing separability measures and the newly developed DCSI on 9 exemplary data sets, as shown in Fig.~\ref{fig:SepExmp}\label{tab:SepExmp}}
\centering
\begin{tabular}[t]{lrrrrrrrrr}
\toprule
  & \textbf{A}& \textbf{B}& \textbf{C}& \textbf{D}& \textbf{E}& \textbf{F}& \textbf{G}& \textbf{H}& \textbf{I}\\
  & dist = 2 & dist = 4 & dist = 8 & outlier & moon & circle & random & lin. sep. & 3 comp.\\
\midrule
\textbf{CVIs:}\\
Dunn* & 0.01 & 0.29 & 0.57 & 0.00 & 0.15 & 0.18 & 0.00 & 0.01 & 0.09\\
CH* & 0.66 & 0.89 & 0.97 & 0.97 & 0.39 & 0.00 & 0.00 & 0.38 & 0.00\\
DB* & 0.61 & 0.77 & 0.87 & 0.86 & 0.46 & 0.05 & 0.02 & 0.46 & 0.00\\
Sil* & 0.78 & 0.89 & 0.94 & 0.94 & 0.67 & 0.58 & 0.50 & 0.68 & 0.68\\
CVNN* & 0.61 & 0.74 & 0.83 & 0.83 & 0.57 & 0.52 & 0.40 & 0.56 & 0.59\\
\addlinespace
\multicolumn{3}{l}{\textbf{Distributional:}}\\
DSI & 0.70 & 0.99 & 1.00 & 1.00 & 0.36 & 0.58 & 0.01 & 0.44 & 0.75\\
\addlinespace
\multicolumn{4}{l}{\textbf{Neighborhood-based:}}\\
N1 & 0.96 & 1.00 & 1.00 & 0.99 & 1.00 & 1.00 & 0.31 & 0.98 & 1.00\\
N2 & 0.88 & 0.97 & 0.98 & 0.95 & 0.97 & 0.97 & 0.50 & 0.90 & 0.98\\
N3 & 0.97 & 1.00 & 1.00 & 1.00 & 1.00 & 1.00 & 0.52 & 0.99 & 1.00\\
LSC & 0.15 & 0.43 & 0.50 & 0.34 & 0.17 & 0.15 & 0.00 & 0.13 & 0.33\\
\addlinespace
\multicolumn{3}{l}{\textbf{Graph-based:}}\\
Density & 0.17 & 0.19 & 0.19 & 0.18 & 0.15 & 0.13 & 0.09 & 0.15 & 0.19\\
ClsCoef & 0.67 & 0.70 & 0.73 & 0.73 & 0.78 & 0.75 & 0.62 & 0.68 & 0.72\\

\addlinespace
\multicolumn{3}{l}{\textbf{MST-based:}}\\
 $\text{DBCV}^*_{\text{all}}$  & 0.15 & 0.96 & 1.00 & 1.00 & 0.98 & 0.99 & 0.01 & 0.06 & 0.36 \\ 
  $\text{DBCV}^*_{\text{core}}$ & 0.70 & 1.00 & 1.00 & 1.00 & 0.98 & 0.98 & 0.02 & 0.08 & 0.37 \\ 
\textbf{DCSI} (ours) & 0.39 & 0.91 & 0.93 & 0.93 & 0.85 & 0.84 & 0.01 & 0.23 & 0.27\\
\bottomrule
\end{tabular}
\end{table}

\noindent Figure \ref{fig:SepExmp} and Table \ref{tab:SepExmp} show 9 simulated data sets and the evaluation of the presented separability measures.
These example data sets aim to illustrate the problems of existing separability measures described above.
\textbf{A}, \textbf{B} and \textbf{C} are drawn from mixtures of two Gaussians with varying distance of means ($2, 4, 8$). 
These data sets are used to investigate the sensitivity of the presented measures with regard to the distance of components. 
\textbf{D} shows the same data as \textbf {C}, but one outlier (red point) is added.
\textbf{E} and \textbf{F} depict classes of non-spherical shape.
The data in \textbf{G} is drawn from one Gaussian and the labels are assigned randomly, so it should be considered the least separable.
\textbf{H} and \textbf{I} reflect the idea that a separability measure for clustering should behave differently from a measure for classification (similar to Figure \ref{fig:ClusterVsClass}, see the explanations in Section \ref{sec:sepClust}).

DBCV was both calculated on all points as originally proposed in \citet{DBCV} as well as using only core points, similar to DCSI.

These examples demonstrate the already mentioned properties and disadvantages of previously described separability measures:
CVIs (first five rows) yield low values for very well-separated clusters with complex shapes like in data sets E and F, but mostly capture the lack of cluster separability of data sets like H and I.
Complexity measures (N1 to ClsCoef), in contrast, do not favor classes of a certain (e.g., spherical) shape, but mostly yield (too) high values for data sets like H and I.

DCSI is able to overcome the disadvantages of the existing measures:
Unlike most other measures except LSC or Density, the DCSI of touching, but not strongly overlapping classes (data set A) is low but not close to zero.
Unlike LSC or Dunn*, DCSI of compact and distinct classes is high and increases with distance, but only up to the distance relevant for separability ($\text{A} < \text{B}\approx \text{C}$). Unlike Dunn* or LSC, DCSI is robust to outliers (D). 
Unlike most CVIs, DSI or LSC, DCSI correctly assigns high separability even if clearly separated classes have complicated shapes (E, F) and also correctly assigns zero separability to random data (G), unlike Sil*, CVNN*, N1, N2, N3 and ClsCoef. 
Unlike N1, N2, N3 and some CVIs, DCSI of data sets whose class labels do not correspond to connected components is relatively low (H, I). 

 \section{Experiments}
\label{sec:exp}
In this section, the results of extensive experiments on synthetic and real-world data are reported. 
The aim is to investigate the behavior of the presented measures and their ability to quantify separability in different situations in more detail.

The separability measures are also calculated on two- or three-dimensional embeddings obtained from the manifold learning algorithm \emph{UMAP} \citep[\emph{Uniform Manifold Approximation and Projection},][]{mcinnes}.
\citet{herrmann} show both from a practical and theoretical perspective that UMAP can considerably improve the performance of DBSCAN by amplifying the distinction between dense and sparse regions.
It is therefore also of interest to evaluate the separability measures on UMAP embeddings.

\subsection{Results on synthetic data sets}
\label{sec:expSynth}

\paragraph{Data sets and procedure:}
Nine experiments on two-class synthetic data were conducted. 
The nine different settings encompass a variety of difficulties for separability measures, such as:
clusters of different density, clusters of non-convex shape such as nested circles, moons and intertwined spirals as well as high-dimensional data sets with many irrelevant features or nested $n$-spheres.

For each of the nine settings, a large number of different data sets are created by varying parameters relevant for separability such as the distance of the classes or the variance of the noise, for a total of 6298 data sets overall. This allows for a thorough investigation of the sensitivity of the separability measures with regard to the parameters of the data sets.
Details on the parameters and their ranges for the nine different settings can be found in Appendix \ref{App:Synth}.

Note that the data sets are \emph{not} Monte Carlo samples, i.e., from each data generating process (DGP), only one data set is drawn.
As outlined before, we consider separability a property of a specific data set and the conducted experiments are more relevant to assess this than Monte-Carlo experiments.
We conducted an additional experiment on the variance of DCSI when evaluated on data sets that are sampled from the same DGP.
The detailed results can be found in Appendix \ref{sec:AppVar}. 
The experiment showed that when evaluated on Monte Carlo samples, the variance of DCSI is small both when the data set is clearly separable or clearly not separable. 
For DGPs which can both lead to realizations with two distinct components or touching components, the variance of DCSI is high.
The clustering performance of DBSCAN shows the same behavior, which motivates our view of separability being a property of a data set rather than a property of the underlying DGP.

For each data set, all 15 separability measures are calculated both on the raw data and their 2D UMAP embeddings.
Furthermore, DBSCAN is applied to both the raw data and the embeddings with $\varepsilon \in [0.01, 10]$ ($\varepsilon \in [0.01, 50]$ for higher dimensional data) and a step size of $0.01$.
The resulting clustering for each $\varepsilon$ is then evaluated using the \textit{Adjusted Rand Index} (ARI) \citep{ARI}.
ARI measures the similarity between the clustering solution and the true labels.
We then use the maximum ARI (i.e., $\max_{\varepsilon} ARI(\varepsilon)$) as a measure for the performance of DBSCAN on this data set.

In order to explore the connection between the performance of DBSCAN and the different aspects of separability quantified by the presented measures, we compute the correlations between the separability measures and the (maximum) ARI.
A high value of ARI is achieved if the clustering solution is similar to the true labels, i.e., if DBSCAN is able to detect the ``correct'' classes (induced by the given labels): the data set's separability is high. This should also be indicated by the separability measures, so higher correlations of a separability measure with ARI are more desirable.

We present a selection of the most relevant findings here, additional figures are shown in Appendix \ref{App:Synth}.
Each of the 6298 data sets represents one observation.

\begin{figure}
\begin{center}
\includegraphics[width = \textwidth]{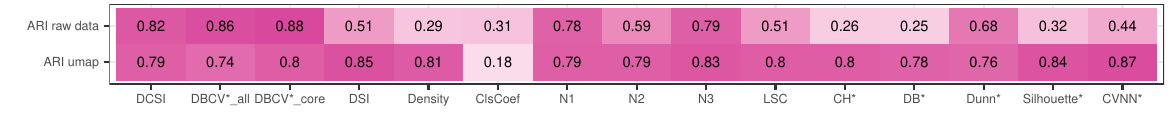}
\end{center}
\caption{Spearman correlation of separability measures and ARI for all 6298 synthetic data sets. See the text (``Overall results'') and the caption of Figure \ref{fig:corApp} for more details.\label{fig:cor}}
\end{figure}

\begin{figure}
\begin{center}
\includegraphics[height = 6.8in, width=4.8in]{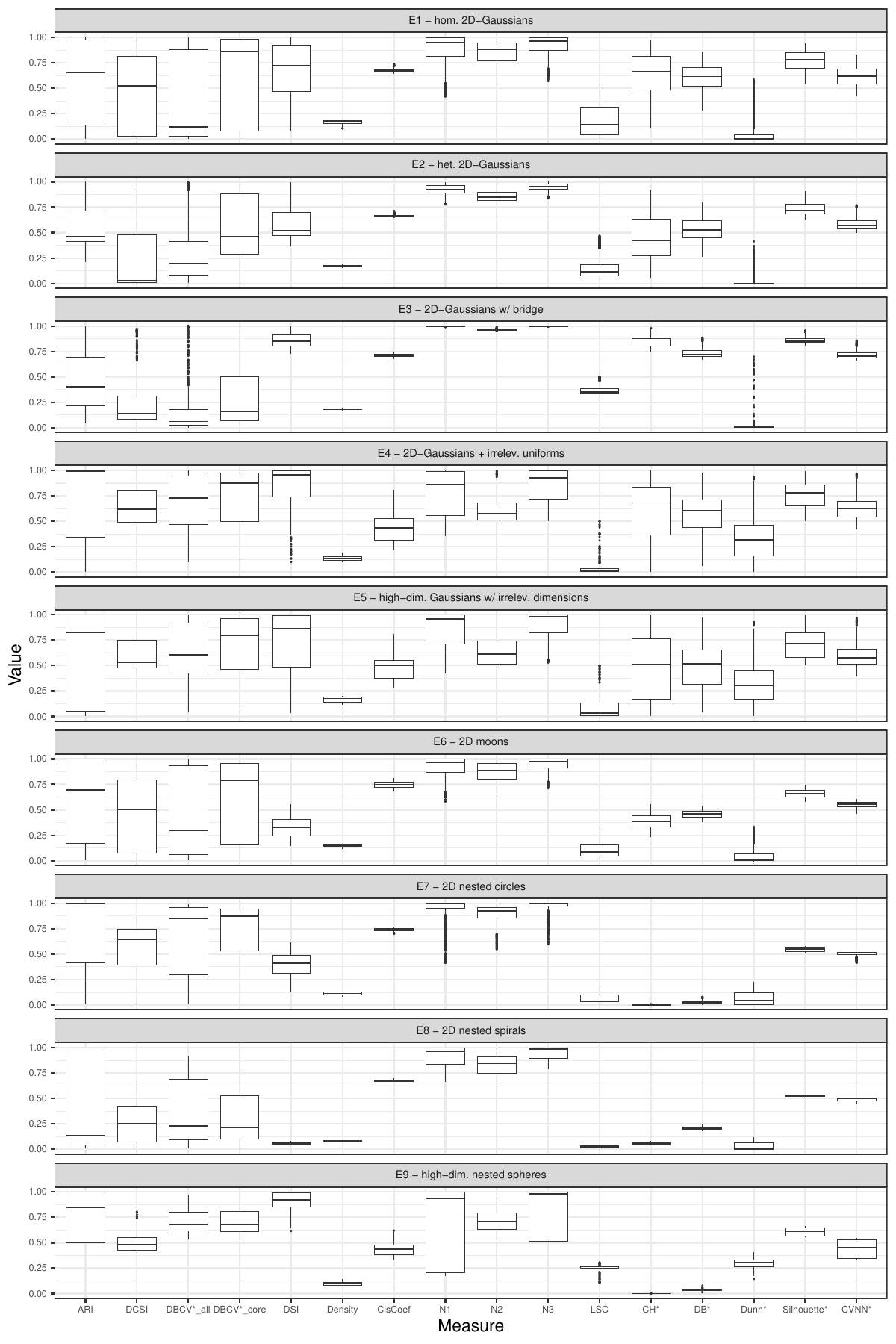}
\end{center}
\vspace*{-4mm}
\caption{Synthetic experiments: Boxplots of separability measures and ARI on raw data.
For each experiment E1-E9, several data sets were generated by varying the parameters (e.g., different distances between classes and different noise variances yield 1519 data sets for E1). For each data set, 15 separability measures and ARI are calculated and the resulting values are shown as boxplots in order to investigate the different ranges across the experiments and the separability measures. 
The most important findings are summarized in the text. \label{fig:BoxplSynth}
}
\end{figure}

\begin{figure}
\begin{center}
\includegraphics[height = 6.8in, width=4.8in]{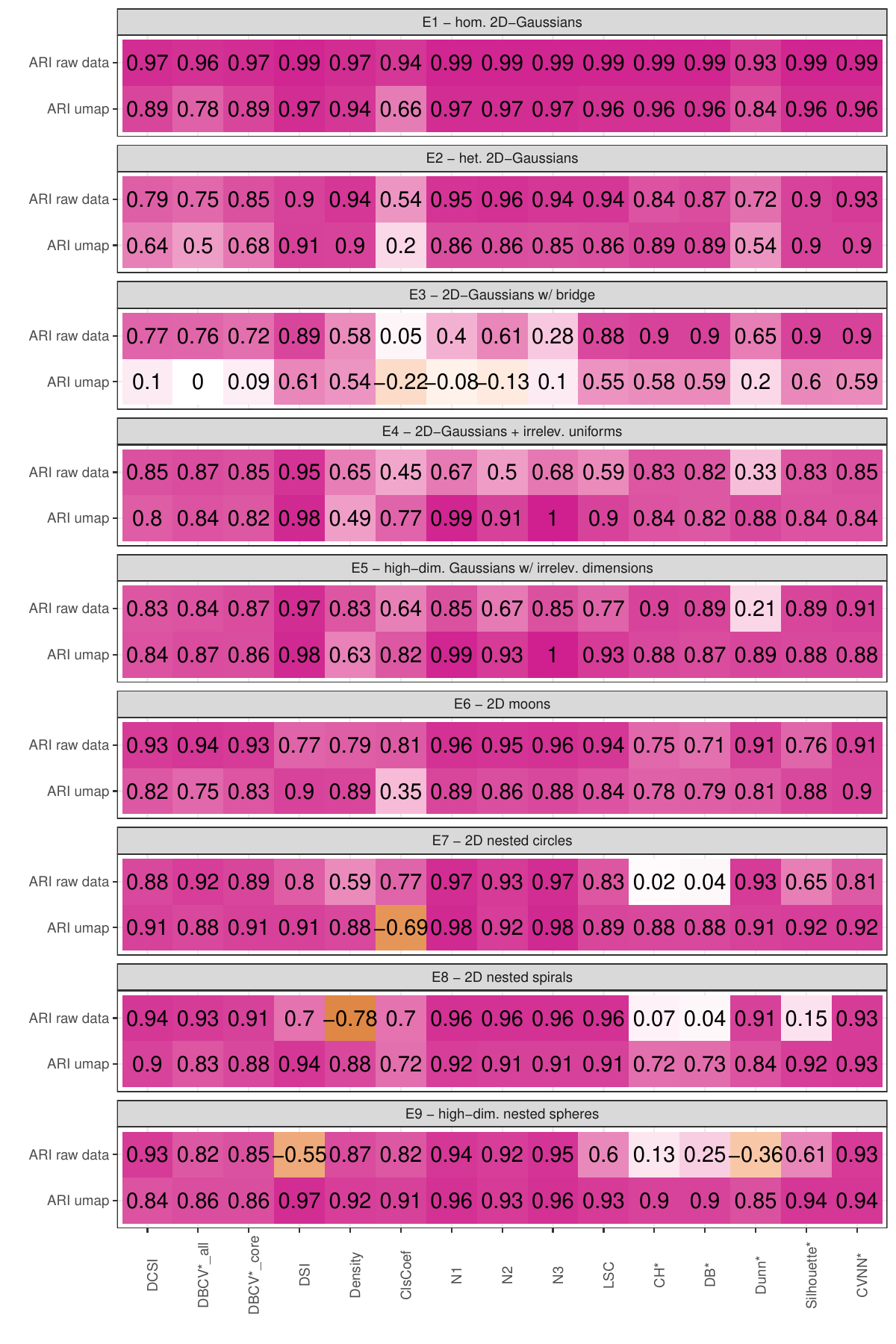}
\end{center}
\vspace*{-9mm}
\caption{Synthetic experiments: Spearman correlations of separability measures and ARI grouped by the nine experiments.
For each data set, 15 separability measures and ARI are calculated on both the raw data and a 2D UMAP embedding.
The correlations between ARI and the separability measures are shown for the nine experiments separately.
A high correlation is desirable, as ARI measures the performance of DBSCAN and thereby indicates if a data set is easy to cluster (i.e., has a high separability), which should also be reflected by the separability measures. 
The most important findings are summarized in the text.
 \label{fig:corApp}}
\end{figure}

\paragraph{Overall results:}
For each of the 6298 data sets, one obtains values of the 15 separability measures and maximum ARI for both the raw data and the 2D UMAP embedding.
Figure \ref{fig:cor} shows the Spearman rank correlations of the separability measures with maximum ARI on the raw data and the embeddings.
In Figure \ref{fig:BoxplSynth} and \ref{fig:corApp}, the results are grouped by the nine experiments in order to obtain deeper insights.
Figure \ref{fig:BoxplSynth} shows boxplots of the values of all separability measures and maximum ARI in order to compare the different ranges.
In Figure \ref{fig:corApp}, the Spearman correlations of the measures with ARI both on the raw data and the UMAP embeddings are shown, again grouped by the nine experiments.  
See \citet{MA} for additional results.

The correlations on the raw data in Figure \ref{fig:cor} are lower than most observed correlations for the separate experiments (e.g., for DSI and N2, see Figure \ref{fig:corApp}).
This might be due to the different ranges for different experiments: 
DSI for example highly correlates  with ARI for both experiment 1 and 7 (Figure \ref{fig:corApp}), but has much smaller values for the nested circles in experiment 7 than for the two-dimensional Gaussians in experiment 1, while ARI takes values across the whole range for both experiments (Figure \ref{fig:BoxplSynth}).

DCSI has the highest correlation with ARI of all separability measures on the raw data.
This indicates that DCSI is able to quantify separability in different settings comparatively well, independent of the shape of the classes or other characteristics of the specific data set.
Similar to some other measures with high correlations (N1, N3), DCSI does not favor classes of a certain shape.
$\text{CH}^*$ and $\text{DB}^*$ on the other hand cannot adequately measure separability on classes of arbitrary shape (e.g., nested circles), which is indicated by the lowest correlations with ARI of all measures (on the raw data).

The correlations of almost all measures are higher on the UMAP embeddings than on the raw data.
Since UMAP tends to yield embeddings with compact, spherical clusters that are not intertwined, the embeddings are much less diverse (e.g., Figure \ref{fig:emb_synth} in the Appendix) than the original data and this is likely to increase the correlation with ARI.

\paragraph{Weaknesses of existing measures:}
Most of the separability measures have a high correlation with ARI both on the raw data and the UMAP embeddings.
However, the synthetic experiments confirm the disadvantages of some existing measures mentioned in Section \ref{sec:ExistSep}:
Most CVIs, especially $\text{CH}^*$ and $\text{DB}^*$, are not suitable for clusters of arbitrary shape, see the low correlations with ARI (raw data) for experiments 7 and 8 (nested circles and spirals) in Figure \ref{fig:corApp} and the low values for all data sets of these experiments in Figure \ref{fig:BoxplSynth}.
DSI also has some difficulties with non-convex clusters, as the values  for experiments 6,7 and 8 (nested moons, circles and spirals) are much smaller than those of the first five experiments (Figure \ref{fig:BoxplSynth}).

The complexity measures, and the neighborhood measures in particular, have low correlations with ARI for touching classes (experiment 3, Gaussians with bridge, Figure \ref{fig:corApp}). 
As most of the data sets in experiment 3 are linearly separable, the classification complexity is low (so the values of N1, N2 and N3 are very high, see Figure \ref{fig:BoxplSynth}), but the classes cannot be seen as two density-based clusters if they touch.

\paragraph{DCSI lacks robustness against unsuitable embeddings:}
As Figure \ref{fig:BoxplSynth} shows, the values of DCSI have a wide range for most experiments and the correlations with ARI are relatively high (Figure \ref{fig:corApp}).
However, because of its definition using the minimum distance between core points of different classes, DCSI can drop sharply if UMAP merges a group of points to the wrong class.
Two data sets where this is the case are shown in Figure \ref{fig:emb_synth} in the Appendix.
This explains the low correlation of DCSI and ARI on the UMAP embeddings for experiment 3, as this situation often occurs when the clusters in the original data slightly touch.
Choosing a higher \textit{MinPts}-value would mitigate this effect. 
See Appendix \ref{App:RW:MinP} for more information on the effect of \textit{MinPts} on a real world data set.

\paragraph{High-dimensional data sets - curse of dimensionality:}
The high dimensionality of the data sets in experiments 4, 5 and 9 lead to interesting effects for some separability measures:
Many measures compare within- and between-cluster distances. 
As irrelevant dimensions are added (experiments 4 and 5), the pairwise distances increase and the intra- and inter-cluster distances become more similar. 
This leads to relatively low correlations of the neighborhood measures and $\text{Dunn}^*$ with ARI for these two experiments (Figure \ref{fig:corApp}).
This effect also explains why DCSI has values close to $0.5$ for data sets with many irrelevant features, although the data is not separable (see Table \ref{tab:e4} in the appendix).

Other interesting effects occur for the high-dimensional nested spheres in experiment 9.
As the same amount of points is sampled from both spheres, the density of the inner sphere is higher and it is always possible for DBSCAN to correctly detect the inner sphere as a cluster and classify the outer sphere as noise points, so the smallest values of maximum ARI are $0.5$ (Figure \ref{fig:BoxplSynth}).
DSI is highly correlated with ARI for all experiments except experiment 9.
Figure \ref{fig:e9} in the Appendix shows the intra- and between-class distances (ICD and BCD) for 2-spheres and 1000-spheres.
As the dimension increases, the variance of the distances decreases, so the distributions of ICD and BCD are less similar, which leads to a higher DSI for high-dimensional spheres.
ARI on the other hand decreases as the dimension increases.

These effects show that one should be careful when separability measures are applied to (intrinsic or artificially) high-dimensional data.
 \subsection{Results on real-world data sets}
\label{sec:RW}

\paragraph{Data sets:}

In order to investigate their behavior on some frequently used data sets, DCSI and the other separability measures were evaluated on the label sets of MNIST and fashion MNIST (FMNIST, both the original 10-class and a 5-class version) and their 3-dimensional UMAP embeddings.
Additionally, the separability measures are not only calculated for the whole data set but also for each pair of classes.
The characteristics of these data sets can be found in Table \ref{tab:dataRW}.
For all data sets, a subsample was drawn for computational reasons.

\begin{table}
\caption{Characteristics of real data sets: number of observations $n_{obs}$ (subsample), original size ($n_{orig}$), number of classes $n_c$, number of features $p$. \label{tab:dataRW}}
\begin{center}
\begin{tabular}[t]{|p{43mm}|p{19mm}|l|l|p{55mm}|}
\hline
Name & $n_{obs}$ ($n_{orig}$) & $n_c$ & $p$ & Description \\
\hline
MNIST \newline \citep{mnist} & 10000  \newline (70000) & 10 & 784 & Handwritten digits, \newline 28x28 grayscale images\\
\hline
FMNIST-10 \newline \citep{fmnist} & 10000 \newline (70000) & 10 & 784 & Fashion products of 10 classes, \newline 28x28 grayscale images\\
\hline
FMNIST-5 \newline \citep{Mukh} & 10000 \newline(70000) & 5 & 784 & 5-class version of FMNIST-10 \\
\hline
\end{tabular}
\end{center}
\end{table}

Note that the subsample for FMNIST-10 and -5 is the same, so the clustering is only computed once and evaluated for both label sets.
The classes in FMNIST-10 are: 0 = T-Shirt/Top, 1 = Trouser, 2 = Pullover, 3 = Dress, 4 = Coat, 5 = Sandal, 6 = Shirt, 7 = Sneaker, 8 = Bag, 9 = Ankle boot.
The classes in FMNIST-5 are: 1 = T-Shirt/Top, Dress, 2 = Trouser, 3 = Pullover, Coat, Shirt, 4 = Bag, 5 = Sandal, Sneaker, Ankle Boot.

Details on the choice of parameters and the selection of the separability measures shown in this section can be found in Appendix \ref{App:RW}.

\paragraph{General results:} 
The results of mean pairwise DCSI with $\textit{MinPts}=50$ and a selection of other well-performing separability measures are summarized in Table \ref{tab:resRW}.
Further options to define a multi-class version of DCSI are evaluated and discussed in Appendix \ref{sec:AppMultiClass}.

All separability measures indicate that UMAP improves the separability.
This is in line with the clustering results (column ``max ARI''),  which makes the separability measures a useful tool to evaluate the quality of higher dimensional UMAP embeddings. 
However, only N2 already indicates on the raw data that FMNIST-5 is easier to cluster than FMNIST-10.
All measures except CH correctly assign a higher separability to MNIST and FMNIST-5 than to FMNIST-10.
\begin{table}
\caption{Results on real-world data: maximum ARI and selected separability measures (3D UMAP embeddings) \label{tab:resRW}}
\begin{center}
\begin{tabular}[t]{|l|l|r|r|r|r|r|}
\hline
Data & Embedding & max ARI & DCSI & DSI & N2 & $\text{CH}^*$ \\
\hline
MNIST & Raw & 0.10 & 0.60 & 0.35 & 0.60 & 0.21\\

MNIST & UMAP & 0.77 & 0.93 & 0.82 & 0.76 & 0.89\\
\hline
FMNIST-5 & Raw & 0.10 & 0.56& 0.43 & 0.62 & 0.31\\

FMNIST-5 & UMAP & 0.76 & 0.78 & 0.79 & 0.80 & 0.82\\
\hline
FMNIST-10 & Raw & 0.07 & 0.57 & 0.47 & 0.56 & 0.40\\

FMNIST-10 & UMAP & 0.41 & 0.73 & 0.72 & 0.66 & 0.87\\
\hline
\end{tabular}
\end{center}
\end{table}

\paragraph{Pairwise separability:}

\begin{figure}
\begin{center}
\includegraphics[height = 6.8in, width = 4.8in]{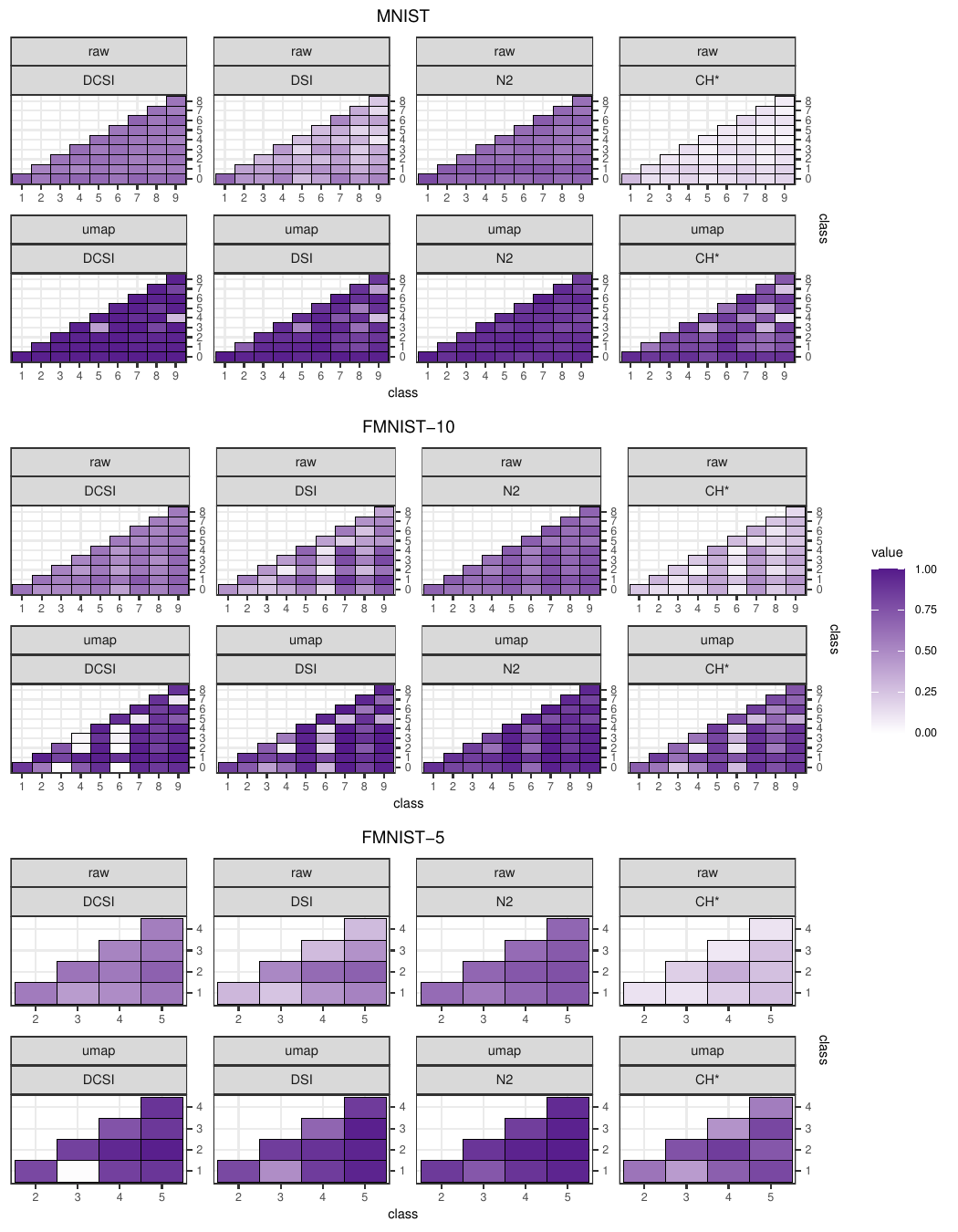}
\end{center}
\caption{Pairwise separability of MNIST, FMNIST-10 and -5 (3D UMAP embeddings)\label{fig:heat}}
\end{figure}

\begin{figure}
\begin{center}
\includegraphics[width =5.5in]{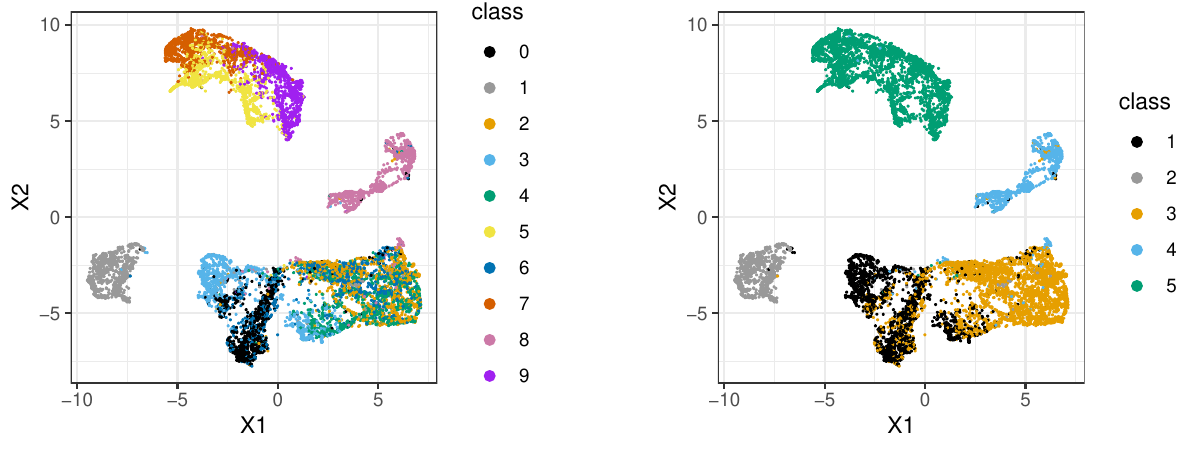}
\end{center}
\caption{FMNIST-10 and -5, 2D UMAP embedding \label{fig:fmnist2D}}
\end{figure}

\begin{figure}
\begin{center}
\includegraphics{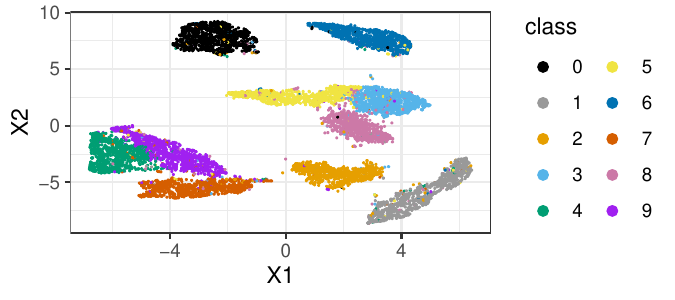}
\end{center}
\caption{MNIST, 2D UMAP embedding \label{fig:mnist2D}}
\end{figure}

One possible application of separability measures is to identify pairs of classes that are not clearly separable and might therefore not be suitable for the evaluation of hard clustering algorithms. The separability for all pairs of classes is shown in Figure \ref{fig:heat}.

The top row of each of the three plots (pairwise separability on raw data) shows that for most measures, the variance between the pairs of classes is relatively low.
For DCSI for example, most values are close to $0.5$, which might be due to the high dimensionality of the data sets: 
As already mentioned in Section \ref{sec:expSynth}, as the dimension increases, the pairwise distances become larger and differ less between the classes, which leads to similar values for separation and connectedness and therefore a DCSI close to $0.5$.
Many other measures also rely on the distinctness of distances between points of different classes, so separability values for high-dimensional data should be handled with care and this section rather focuses on the results on UMAP embeddings.

A comparison of the separability of the UMAP embeddings (3D) and the visualizations of 2D embeddings in Figures \ref{fig:fmnist2D} and \ref{fig:mnist2D} shows that DCSI correctly identifies touching or overlapping classes.
This is an advantage compared to the other measures, which mainly quantify separability from a classification-based point of view.
For example, DCSI is the only measure that indicates that classes 7 and 9 in FMNIST-10 (sneaker and ankle boot) slightly touch (see Figure \ref{fig:fmnist2D}) and therefore do not form meaningful clusters.
The same applies to classes 1 and 3 in FMNIST-5.

N2 seems to be a good indicator for the overall difficulty of the data sets (see Table \ref{tab:resRW}), but it fails to clearly differentiate between pairs of classes that are clearly or barely separable.
This can be explained by its definition: 
If two classes touch or overlap, the specific value of N2 is very sensitive to the amount of points far away from the border, which leads to a high separability for touching pairs of classes like 4 and 9 in MNIST (Figure \ref{fig:mnist2D}).
While this behavior is appropriate if separability is measured from the perspective of classification, it is not desirable for a clustering-based view. 

 In summary, these results emphasize the ability of DCSI to identify (pairs of) classes that might be separable by a suitable classifier but do not correspond to meaningful (density-based) clusters.

 \section{Discussion \& Conclusion}
\label{sec:disc}

Our review in Section \ref{sec:ExistSep} shows that existing measures of separability only each cover some aspects of separability and that no measure is able to incorporate all aspects necessary to quantify the separability of density-based clusters.
Most complexity measures and DSI focus on classification, so they do not measure connectedness but only between-class separation.
Most cluster validity indices (CVIs) on the other hand favor clusters of spherical shape as they take compactness of classes into account.
In order to overcome some disadvantages of the existing measures, we propose a new measure of separability, DCSI, which quantifies both within-class connectedness and between-class separation in a way that is suitable for density-based clustering.

Extensive experiments on synthetic data show that DCSI correlates highly with the clustering performance (measured by DBSCAN's maximally achieved ARI) in almost all settings.
Additionally, DCSI has the highest correlation with ARI of all presented separability measures if all synthetic data sets are evaluated jointly.
Our results also indicate that DCSI can lack robustness if its \textit{MinPts} parameter is too small and that it is less discriminatory in high-dimensional data, similar to other separability measures that rely on the distinctness of pairwise distances.

The results on real-world data show that separability measures are a useful tool for the evaluation of UMAP embeddings with more than two dimensions, especially if higher values of \textit{MinPts} are used for increased robustness.
Furthermore, DCSI is a valuable complement to existing measures such as neighborhood-based measures, especially for the quantification of pairwise separability:
DCSI can detect overlapping or touching classes and therefore identify classes that do not form meaningful density-based clusters.

Our results also support the importance of issues raised in \citet{herrmann} and \citet{schubert2017}:
Does it make sense to evaluate clustering algorithms using labeled data without knowing if the given classes correspond to meaningful clusters? 
Separability measures might be a useful tool to identify suitable data sets for methodological research.
Similar to clustering algorithms, each separability measure implicitly defines its own truth of ``meaningful'' clusters and DCSI is suited particularly well for density-based clustering.
In applied research, DCSI can be used as a CVI in order to evaluate the quality of a given clustering and choose the parameters of DBSCAN, especially $\varepsilon$. 

The experiments have shown that the choice of \textit{MinPts} can strongly affect the separability, as it determines which groups of points are considered core points.
The effects of the choice of \textit{MinPts} need further investigation.
Similarly, the sensitivity of DCSI with regard to $\varepsilon_i$ and how it can be chosen in a way that is ``optimal'' remains an open question.

The high correlation of DCSI and (maximum) ARI indicates that it might be possible to predict the (maximum) ARI of a data set based on the separability measures.
Another interesting question is if it is possible to identify certain types or classes of problems based on the separability measures by investigating the distribution of problems in the multidimensional space spanned by the separability measures, similar to \citet[Chapters 4-6]{Ho2002}.

\pagebreak

\begin{center}
{\large\bf Compliance with Ethical Standards}
\end{center}

  This work has been funded by the German Federal Ministry of Education and Research and the Bavarian State Ministry for Science and the Arts. 
  The authors of this work take full responsibility for its content.
  The authors have no competing interests to declare.

\paragraph{Data availability}
All real-world data sets can be downloaded from the github repository mentioned below.
The simulated data sets can be reproduced with the provided code.

\paragraph{Code availability}
The code and data to reproduce the results can be found on Github: \url{https://github.com/JanaGauss/dcsi}.
All analyses were conducted in \emph{R} \citep{R}.
The complexity measures are computed with the \emph{ECol} package\citep{ECol} and all CVIs except CVNN with the \emph{clusterCrit} package \citep{RClust}.
CVNN, DSI and DCSI are calculated using the first author's implementations.
The packages used for DBSCAN and UMAP are \emph{dbscan} \citep{hahsler2019} and \emph{umap} \citep{Rumap}.

\newpage
\bibliographystyle{jabes}

\bibliography{references}

\pagebreak
\appendix 
\begin{center}
{\large\bf APPENDIX}
\end{center}

\section{Variance of DCSI}
\label{sec:AppVar}

\begin{figure}
\begin{center}
\includegraphics[width = 4in]{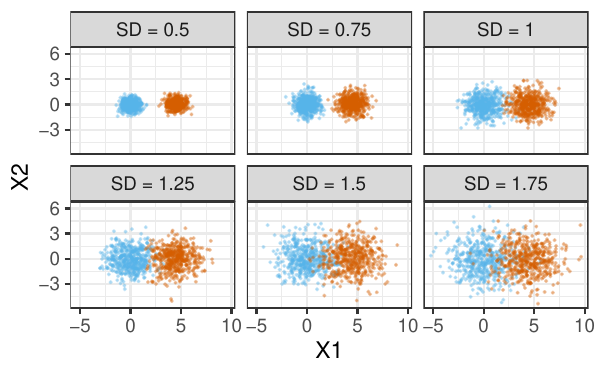}
\end{center}
\caption{Exemplary data sets with varying standard deviation.\label{fig:datVar}}
\end{figure}

In order to exemplarily investigate the variance of DCSI on data sets drawn from the same data generating process (DGP), DCSI was evaluated on 200 data sets each from seven different DGPs: two two-dimensional Gaussians with mean $(0,0)$ and $(4.5, 0)$ and covariance $\sigma^2 I_2$, where $\sigma = 0.5, 0.75., \ldots, 2$. 
From both Gaussians, $n_1 = n_2 = 500$ data points were drawn.
Exemplary data sets are shown in Figure \ref{fig:datVar}.

\begin{figure}
\begin{center}
\includegraphics[width = 4in]{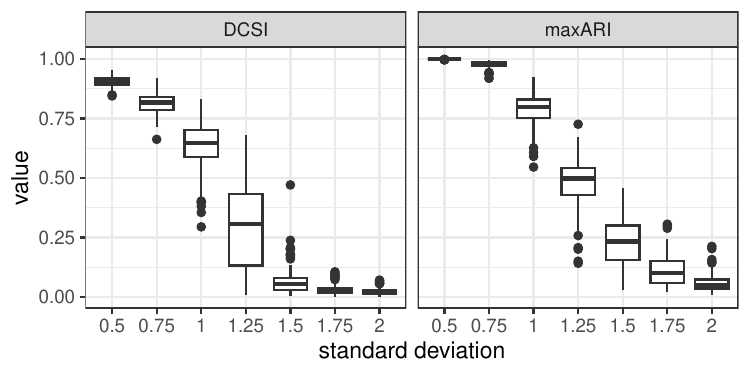}
\end{center}
\caption{Results of DCSI and maximum ARI for seven DGPs (two-dimensional Gaussians with varying standard deviation). From each DGP, 200 data sets of size $n=1000$ were drawn.\label{fig:VarBox}}
\end{figure}

Additionally, DBSCAN was evaluated on each data set with different values of $\varepsilon$ and maximum ARI is taken as a measure of difficulty of the respective clustering task.
The results are shown Figure \ref{fig:VarBox}.

The variance of DCSI is small both for data sets that are clearly separable and clearly not seperable (see Fig.~\ref{fig:datVar}), which is in line with the clustering performance.
For the ``intermediate'' data sets, the variance of both DCSI and maximum ARI is high, as for these data sets, the concrete realization determines if the data set is separable in two distinct components or if these components touch.
We argue that these results show that DCSI (and separability measures in general) should not be seen as a metric that measures a property of a DGP but rather a property of a data set.
The focus of the experiments in Section \ref{sec:expSynth} is therefore not on estimating the average DCSI and its variance for certain DGPs but rather on the relation between DCSI and the clustering performance measured by maximum ARI for data sets from different DGPs.
For the 1400 data sets in Fig.~\ref{fig:VarBox}, the Spearman correlation between DCSI and maximum ARI is $0.94$.
In order to investigate the strength of the correlation between DCSI and the clustering performance as the parameters of the underlying DGP changes, for the experiments in Section \ref{sec:expSynth}, only one data set is drawn from each DGP, which allows to sample data sets from a dense grid of parameters.

\section{Experiments on synthetic data and additional plots}
\label{App:Synth}

Nine experiments with 6298 data sets in total were conducted for Section \ref{sec:expSynth}.
Each data set consists of two classes with $n_1 = n_2 = 500$ (except for experiment 3) that are sampled from two (more or less separated) components.
For each combination of parameters, there is one data set (e.g., for experiment 1, there are 49 values for $d$ and 31 for $\sigma$, so $49\cdot 31=1519$ data sets in total).

\begin{itemize}
\item \textbf{Experiment 1 (homogeneous~(hom.)~2D-Gaussians)}: 
Two two-dimensional Gaussians of varying distance and covariance, 1519 data sets. 
Mean first component: $(0,0)$, mean second component: $(d, 0)$ with $d=2, 2.125, 2.25,\ldots, 7.875, 8$.
Covariance: the same in both components, $\sigma^2 I_2$ with $\sigma = 0.5, 0.55, \ldots, 1.95, 2$.
\item \textbf{Experiment 2 (heterogeneous~(het.)~2D-Gaussians)}: 
Two two-dimensional Gaussians with different densities, 1525 data sets.
Mean first component: $(0,0)$, mean second component: $(d, 0)$ with $d=2, 2.125,\allowbreak 2.25,\allowbreak \ldots,\allowbreak 4.875, 5$.
Covariance first component: $0.5^2 I_2$, covariance second component: $\sigma^2 I_2$ with $\sigma = 0.5, 0.55,\ldots, 3.45, 3$.
\item \textbf{Experiment 3 (2D-Gaussians w/ bridge)}: 
Two two-dimensional Gaussians connected by a bridge, 775 data sets.
Mean first component: $(0,0)$, mean second component: $(d, 0)$ with $d=4,\allowbreak 4.25,\allowbreak \ldots,\allowbreak 9.75,\allowbreak 10$. 
Covariance: the same in both components, $0.5^2I_2$.  
A bridge of points $(X_1, X_2)$ is built between the classes by sampling $X_1$ from a uniform distribution on $[0, d]$ and $X_2$ from $\mathcal{N}(0, \sigma^2)$ with $\sigma$ being $0.2$ of the observed standard deviation of $X_2$. 
To obtain labels for the points on the bridge, each point is added to the closest component.
Density of the bridge: The amount of points sampled for the bridge is $c \cdot n$ with $c=0, 0.05, \ldots, 1.45,1.5$ (and $n=1000$).
\item \textbf{Experiment 4 (2D-Gaussians $+$ irrelevant uniforms)}:
Two two-dimensional Gaussians and additional irrelevant features, 324 data sets. 
Mean first component: $(0,0)$, mean second component: $(d, 0)$ with $d=1.5, 1.75,\ldots, 4.75, 5, 10, 20, 50$.
Covariance: the same in both components, $0.5^2I_2$.
Additionally, $n_{irrev}$ further features are sampled uniformly from $[0,1]$ with $n_{irrev} = 0, 1,\ldots, 9, 10, 15, 20, 50, 100, 500, 1000, \allowbreak 2000$ (i.e., the total number of features is $2+ n_{irrev}$).
\item \textbf{Experiment 5 (high-dim.~Gaussians w/ irrelevant dimensions)}: 
Two multidimensional Gaussians, 288 data sets.
The data is sampled from two $p$-dimensional Gaussian with $p=2, 3,\ldots, 9, 10, 15, 20, 50, 100, 500,\allowbreak 1000,\allowbreak 2000$.
Mean first component: $(0,0, \ldots, 0)$, mean second component: $(d, 0, \ldots, 0)$ with $d=1.5, 1.75,\ldots, 4.75, 5,\allowbreak 10,\allowbreak 20, 50$.
Covariance: the same in both components, $0.5^2I_p$.
\item \textbf{Experiment 6 (2D moons)}:
Two two-dimensional moons, 820 data sets.
The data is sampled uniformly from a (2-D) circle with radius 6 and center $(0,0)$. 
The upper moon is shifted horizontally by 6 units. 
Then, the upper moon is shifted vertically by $6 s$ with $s=0, 0.05, \ldots, 0.9, 0.95$ (i.e., for $s=1$, the moons would touch).
Two-dimensional Gaussian noise is added with covariance $\sigma^2I_2$ with $\sigma = 0, 0.05, \ldots, 1.95, 2$.
\item \textbf{Experiment 7 (2D nested circles)}: 
Two two-dimensional nested circles, 861 data sets.
One component is sampled uniformly from a circle with radius 4, the other uniformly from a circle with radius $r$ with $r=5, 5.125, \ldots, 9.875, 10$. 
The center of both circles is $(0,0)$.
Two-dimensional Gaussian noise is added with covariance $\sigma^2I_2$ with $\sigma = 0, 0.05, \ldots, 0.95, 1$.
\item \textbf{Experiment 8 (2D nested spirals)}: 
Two two-dimensional spirals, 51 data sets.
The data is sampled uniformly from two intertwined (2-D) spirals.
Two-dimensional Gaussian noise is added with covariance $\sigma^2I_2$ with $\sigma = 0, 0.05, \ldots, 2.45, 2.5$.
\item \textbf{Experiment 9 (high-dim.~nested spheres)}: 
 Two nested $n$-spheres, 135 data sets.
 One component is sampled uniformly from a $n$-sphere with radius 4, the other uniformly from a $n$-sphere with radius $r$ with $r=10, 20, 50$. The center of both spheres is $(0,0)$, $n = 2, 3,\ldots, 9, 10, 15, 20, 50,\allowbreak 100, 500, 1000$.
 Two-dimensional Gaussian noise is added with covariance $\sigma^2I_2$ with $\sigma = 0, 0.25, 0.5$.
  \end{itemize}
  \begin{figure}
\begin{center}
\includegraphics[height = 6.8in, width = 4.8in]{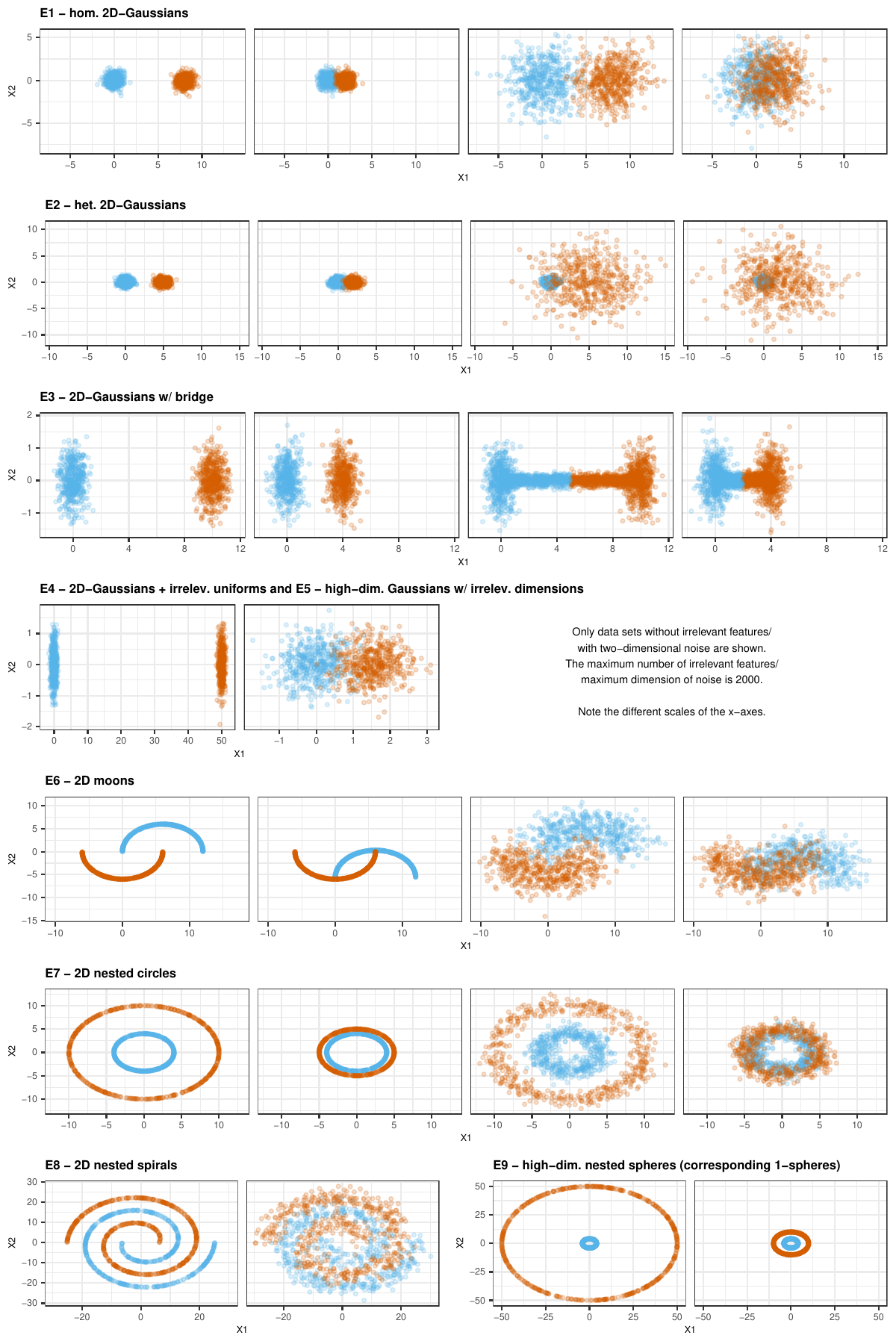}
\end{center}
\caption{Overview synthetic data sets \label{fig:DatSynth}}
\end{figure}

  \begin{figure}
\begin{center}
\includegraphics[width = 4.8in]{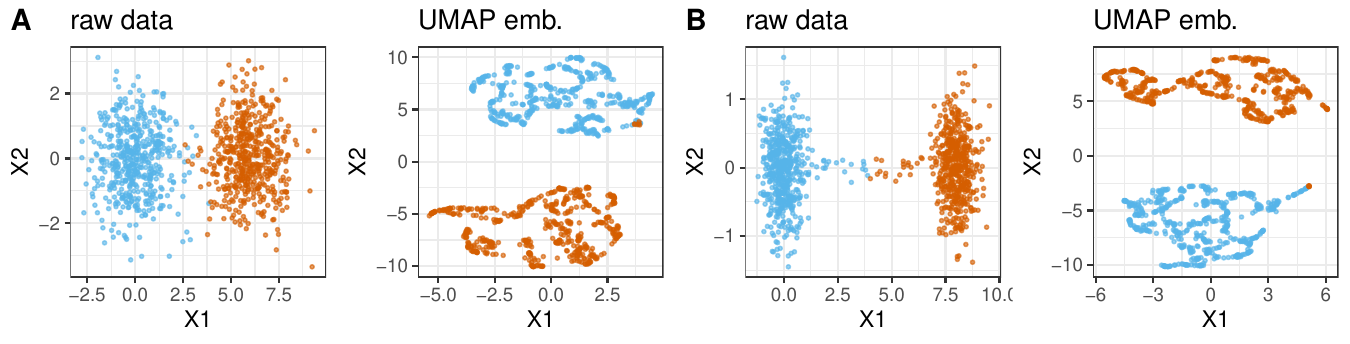}
\end{center}
\caption{Synthetic experiments: Exemplary embeddings where DCSI lacks robustness from experiment 1 (\textbf{A}) and 3 (\textbf{B})\label{fig:emb_synth}}
\end{figure}

 \begin{figure}
\begin{center}
\includegraphics[width = 4.8in]{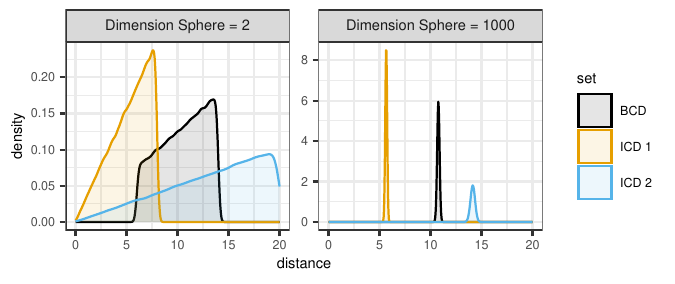}
\end{center}
\caption{Experiment 9: Examples of ICD and BCD sets (DSI) for $r=10, \sigma = 0$\label{fig:e9}}
\end{figure}

E1, E2 and E3 represent different aspects of variation for two-dimensional Gaussians. 
The interesting aspect of E2 is the different density in both components.
E3 aims to answer the questions at which point two components that are (slightly) connected with each other cannot be seen as two clusters anymore.
With E4 and E5, the effect of an artificially high dimension is investigated.
E6, E7, E8 and E9 represent different non-spherical shapes of varying complexity.
E9 is the only setting where the intrinsic dimension of the data is high.

The most ``extreme'' data sets for each experiment are shown in Figure \ref{fig:DatSynth}, e.g., for experiment 1, these are the data sets with $(d,\sigma) \in \{(8,0.5), (2, 0.5), (8, 2), (2,2) \}$ (in this order, the easiest data set in the first column, the most difficult data set in the last one).
For E4 and E5, only the two-dimensional data sets are shown, i.e., the data set without irrelevant features (E4) and with two-dimensional noise (E5). 
As these two-dimensional data sets have the same parameters (same $d$, same covariance) for E4 and E5, only the data sets from E4 are shown.
For E8, only one parameter (the covariance) is varied, so only two data sets are plotted.
For E9, the data set with the lowest dimension is three-dimensional and therefore not shown.
The corresponding two-dimensional data sets (i.e., 1-spheres) with $\sigma = 0$, $r=10$ and 50 are shown instead. 

The parameters were chosen as follows:
\begin{itemize}
\item \textbf{Separability measures}:
The $\varepsilon$-value for the network measures is 0.15 \citep[as in][]{Lorena}, the nearest neighbor parameter for CVNN is $k=10$ \citep[as in][]{LiuEnh}, the \textit{MinPts} parameter for DCSI is 5 and the $\varepsilon_i$ are chosen as proposed in Section \ref{sec:DCSI}.
\item \textbf{DBSCAN}: $\textit{MinPts}=5$, $\varepsilon\in [0.01, 10]$ (and $\varepsilon\in [0.01, 50]$ for high-dimensional data) with a step size of $0.01$.
\item \textbf{UMAP}: UMAP was always used with spectral initialization and $\textit{min-dist}= 0.1$, the nearest neighbor parameter is $k=15$ (this value was chosen based on results of a pilot study).
\end{itemize}

\begin{table}
\caption{Experiment 4: Separation, Connectedness and DCSI on raw data and the embedding for a data set with $d=1.5$ (distance of means) and 2000 irrelevant features. ARI is 0 both on the raw data and the embedding.\label{tab:e4}}
\centering
\begin{tabular}[t]{llrrrrrr}
\hline
$d$ & $n_{irrev}$ & Sep raw & Conn raw & DCSI raw & Sep UMAP & Conn UMAP & DCSI UMAP\\
\hline
1.5 & 2000 & 17.19 & 17.75 & 0.49 & 0.01 & 0.61 & 0.01\\
\hline
\end{tabular}
\end{table}

\section{Experiments on real-world data}
\label{App:RW}

All data sets were standardized (not column-wise but the data was treated as a matrix).
The $\varepsilon$-ranges for DBSCAN are $\varepsilon_{raw} \in [1, 40]$ for the raw data and $\varepsilon_{umap} \in [0.01, 10]$ (MNIST) and $\varepsilon_{umap} \in [0.01, 15]$ (FMNIST-10, -5) for the UMAP embeddings, the step size is $0.01$.
For UMAP, $k=10$ was chosen, as this value yields the best results for most data sets in \citet[Table 5]{herrmann}.
DCSI was calculated with $\textit{MinPts} = 50$, see the investigation on the sensitvity of DCSI to \textit{MinPts} below.
The other parameters are the same as in Section \ref{sec:expSynth}/Appendix \ref{App:Synth}.

Besides DCSI, the results of DSI, N2 and $\text{CH}^*$ are shown in Section \ref{sec:RW}. 
These three measures were selected such that each category presented in Section \ref{sec:ExistSep} has one representative.
N2 and $\text{CH}^*$ were chosen among the complexity measures/CVIs because the values of some other measures with higher correlations with ARI (Figure \ref{fig:cor} \textbf{A}) had almost no variability on the real-world data (N1 and N3 for example had values close to one for most data sets and $\text{Dunn}^*$ was close to zero for almost all embeddings).

\subsection{Sensitivity of DCSI to \textit{MinPts}:}
\label{App:RW:MinP}

As the experiments on synthetic data show, the separability according to DCSI can drop sharply on UMAP embeddings due to a group of points being merged into the ``wrong'' class.
DCSI is based on the maximum and minimum distances of core points, so the definition of a ``core point'' highly affects the separability: 
A core point has at least \textit{MinPts} observation of the same class in its $\varepsilon$-neighborhood.
A small value of \textit{MinPts} thus increases the sensitivity of DCSI to groups of ``outliers'' and DCSI can be robustified by selecting a higher value for \textit{MinPts}.
It might also make sense to choose this parameter based on the group sizes.
However, whether a group of points of a certain size should be considered as ``outliers'' or noise or not, and therefore affect the separability or not, should be determined by the specific application. \\

\begin{figure}
\begin{center}
\includegraphics[width = 4.8in, height = 0.15\textheight]{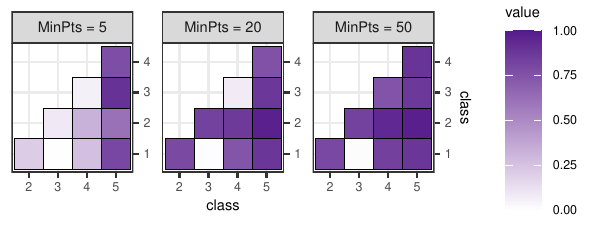}
\end{center}
\caption{Pairwise separability of FMNIST-5 (3D UMAP embeddings) for different \textit{MinPts} values\label{fig:heatR}}
\end{figure}

\noindent In order to exemplarily investigate the sensitivity and behavior of DCSI for different \textit{MinPts} values, the pairwise separability of the UMAP embedding of FMNIST-5 was computed for $\textit{MinPts} = 5, 20, 50$. 
The results are shown in Figure \ref{fig:heatR}.

The low separability of pairs of classes such as 2-3 and 4-3 for $\textit{MinPts}=5$ is caused by single core points that are close to another class.
For $\textit{MinPts}=20$, the separability of 2-3 increases and for $\textit{MinPts} = 50$, the separability of both pairs is relatively high, as certain groups of points are not classified as core points anymore.
Higher values of \textit{MinPts} can therefore enhance the robustness of DCSI to groups of ``outliers''.
At the same time, the separability of the classes 1 and 3  ($1 = \{\text{T-Shirt/Top, Dress}\},
3 = \{\text{Pullover, Coat, Shirt}\}$) is low for all three values of \textit{MinPts}, so DCSI is still able to correctly identify touching classes. 

Similar computations showed that $\textit{MinPts} = 50$ yields meaningful results for FMNIST-10 and MNIST, so $\textit{MinPts} = 50$ was used in Section \ref{sec:RW}.

\subsection{Different multi-class versions of DCSI}
\label{sec:AppMultiClass}

There are different ways to define a multi-class version of DCSI.
Two possibilities suggest themselves: DCSI of a multi-class data set could be defined as some summary of the pairwise DCSIs (group 1) or one could define separation $\Sep_{All}$ and connectedness $\Conn_{All}$ of the entire data set as summaries of separation and connectedness of its classes and set $\DCSI = q/(1+q)$, where $q =  \Sep_{All}/\Conn_{All}$ (group 2).
For the first group, possible options are the mean, median and minimum pairwise DCSI.
For the second group, one could take the mean, median or worst values of separation and connectedness. 
Note that due to its definition, the worst value of intra-class connectness is the highest value of $\Conn$.

In Table \ref{tab:multiclassDCSI}, six different versions are shown with their definition and their evaluation of MNIST and FMNIST.
All values are calculated on 3D UMAP embeddings and using $\textit{MinPts}=50$.

Taking the minimum pairwise DCSI or the worst values of separation and connectedness yields very sensitive measures that rather indicate if there exists a pair of classes that is not well separated, so a perfect clustering is not possible.
The measures in group 2 combine intra-class connectedness and inter-class separation independently, so they ignore the interplay between separation and connectedness of a pair of classes.
The mean or median pairwise DCSI therefore seems more suited to summarize the separability of a multi-class data set.
Since DCSI is bounded between 0 and 1 and it is reasonable to take all values of pairwise separability into account, we suggest that the mean pairwise DCSI is the best way to obtain one value of separability of a multi-class data set.
Furthermore, this is the only measure where the order of the three data sets coincides with their order regarding ARI in Table \ref{tab:multiclassDCSI}.
However, the desired properties of a multi-class DCSI depend on the application and it often makes sense to have a look at different summary statistics of the pairwise DCSIs such as the minimum or certain quantiles.

\begin{table}
\caption{Maximum ARI and different versions of multi-class DCSI evaluated on 3D UMAP embeddings. $\{  \DCSI(C_i, C_j) \}$ denotes $\{  \DCSI(C_i, C_j) , i , j =1,\ldots K, i \neq j\}$, $\{ \Conn(C_i) \}$ denotes $\{ \Conn(C_i), i = 1, \ldots K\}$ etc.\label{tab:multiclassDCSI}}
\begin{center}
\begin{tabular}[t]{|l|l| l |r|r|r|}
  \hline
 \textbf{Group} & \textbf{Measure} & \textbf{Definition} &\textbf{MNIST} &  \textbf{FMNIST-5} & \textbf{FMNIST-10}  \\ 
  \hline
  & max ARI &  &0.77 & 0.76 & 0.41 \\
  \hline
\textbf{1}&mean &$ \mathop{\mathrm{mean}} \{  \DCSI(C_i, C_j) \}  $& 0.93& 0.78 & 0.73  \\ 
   &median&$  \mathop{\mathrm{median}}\{  \DCSI(C_i, C_j)  \}$&0.97 & 0.86 & 0.90  \\ 
   &min &$ \min\{  \DCSI(C_i, C_j)  \} $& 0.29& 0.01 & 0.01  \\ 
   \hline
  \textbf{2} & mean  & \makecell{ $\Sep_{All}  =  \mathop{\mathrm{mean}}\{ \Sep(C_i, C_j)\} ,$ \\ $\Conn_{All}  =  \mathop{\mathrm{mean}}\{ \Conn(C_i)\}$}& 0.97& 0.90 & 0.93  \\ 
   &median & \makecell{ $\Sep_{All}  =  \mathop{\mathrm{median}}\{ \Sep(C_i, C_j)\} ,$ \\ $\Conn_{All}  =  \mathop{\mathrm{median}}\{ \Conn(C_i)\}$}& 0.97& 0.90 & 0.95  \\ 
   & minmax & \makecell{ $\Sep_{All}  =  \min \{ \Sep(C_i, C_j)\} ,$ \\ $\Conn_{All}  =  \max \{ \Conn(C_i)\}$}& 0.29& 0.01 & 0.01  \\ 
   \hline
\end{tabular}
\end{center}
\end{table}

 \section{Definitions of existing separability measures}
\label{App:Def}
As it is desirable that all measures are in \([0, 1]\) (or \([0, 1[\) or \(]0, 1]\) etc.) with \(1\) as best value (highest separability), some measures are slightly modified which is indicated by an asterisk.
The notation is as follows: \(X=x_1, ..., x_n\) is a given data set with \(K\) classes \(C_1, ..., C_K\) of sizes \(n_1, ..., n_K\) with centers \(c_1, ..., c_K\) (i.e., the mean of each class).
\(c\) is the center of the whole data set.
\(d(x, x')\) denotes the Euclidean distance between \(x\) and \(x'\) (unless otherwise stated, see Section \ref{sec:MeasuresGraph}).
For some measures, a distance \(d(C_i, C_j)\) or a similarity \(s(C_i, C_j)\) between two classes \(C_i\) and \(C_j\) is defined.
\(\Sep(X)\) and \(\Comp(X)\) denote index specific definitions of separation and compactness.
For a point \(x_i\), \(y_i\) denotes the class label of \(x_i\).
For more details on the characteristics of the measures, see \citet{MA}.

\subsection{Internal Cluster Validity Indices\label{sec:MeasuresCVIs}}

\paragraph{Dunn Index:}
The Dunn Index \citep{Dunn} is the ratio of separation and compactness, which are defined as follows:
The distance between two classes \(C_j\) and \(C_j\) is the minimum distance between points of these classes.
The separation \(\Sep_{\Dunn}(X)\) of the whole data set \(X\) is given by the minimum distance between two classes \citep{Dunn}.
For a class $C_k$, the diameter \(\operatorname{diam}(C_k)\) is the maximum distance of points in this class.
The compactness \(\Comp_{\Dunn}(X)\) is given by the maximum diameter \citep{Dunn}:
\begin{align*} 
  d(C_i, C_j) & = \min_{x \in C_i, x' \in C_j} d(x,x') ,\\
  \Sep_{\Dunn}(X) &= \min_{i, j = 1, ..., K, i \neq j} d(C_i, C_j) ,\\
  \operatorname{diam}(C_k) &= \max_{x, x' \in C_k} d(x,x') ,\\
  \Comp_{\Dunn}(X) &= \max_{k = 1, ..., K} \operatorname{diam}(C_k).
\end{align*}
The Dunn index is the ratio of \(\Sep_{\Dunn}\) and \(\Comp_{\Dunn}\):
\begin{definition}[Dunn index]
\label{def:unnamed-chunk-42}
\[
\Dunn(X) = \frac{\Sep_{\Dunn}(X)}{\Comp_{\Dunn}(X)}= \frac{ \min_{i, j, i \neq j} (\min_{x \in C_i, x' \in C_j} d(x,x'))}{\max_k (\max_{x, x' \in C_k} d(x,x'))}\] \citep{Dunn}.
As there is no upper limit, the Dunn index is slightly modified to be in \([0,1[\):
\[
\Dunn(X)^* = \frac{\Dunn(X)}{1 + \Dunn(X)}.
\]
\end{definition}

\paragraph{Calinski-Harabasz Index (CH):}
The Calinski-Harabasz Index (CH) \citep{CH} also takes the form \(\Sep_{\CH}/\Comp_{\CH}\) \citep{LiuEnh}.
Separation is measured in terms of the weighted sum of squared distances of the class centers to the center \(c\) of the whole data set.
Compactness is based on the within-group variance \citep{LiuEnh}:
\begin{align*} 
  \Sep_{\CH}(X) &= \frac{1}{K - 1} \sum_{i = 1}^{K} n_i d(c_i, c)^2 ,\\
  \Comp_{\CH}(X) &= \frac{1}{n - K} \sum_{i = 1}^{K} \sum_{x \in C_i} d(x, c_i)^2.
\end{align*}
The CH index is defined as
\begin{definition}[Calinski-Harabasz index]
\label{def:unnamed-chunk-44}
\[
\CH(X) = \frac{\Sep_{\CH}(X)}{\Comp_{\CH}(X)}= \frac{\sum_{i = 1}^{K} n_i d(c_i, c)^2 / (K - 1)}{\sum_{i = 1}^{K} \sum_{x \in C_i} d(x, c_i)^2 / (n - K)} = \frac{n - K}{K - 1} \frac{\sum_{i = 1}^{K} n_i d(c_i, c)^2)}{\sum_{i = 1}^{K} \sum_{x \in C_i} d(x, c_i)^2}
\] \citep{ LiuEnh}.
CH can take arbitrary high values.
The modified version of the CH index is given by
\[
 \CH(X)^* = \frac{\CH(X)^{**}}{1 + \CH(X)^{**}}
\text{ where }\CH(X)^{**} = \frac{K-1}{n-K}\CH(X).
\]
\end{definition}
As this index is used as a CVI and the term \(\sum_{i = 1}^{K} \sum_{x \in C_i} d(x, c_i)^2\) (typically) becomes smaller as the number \(K\) of clusters increases, it is corrected by multiplying with \((n-K)/(K-1)\), which decreases as \(K\) increases.
However, when used as a separability measure, no correction for the number of classes is needed.

\paragraph{Davies-Bouldin Index (DB):}
The Davies-Bouldin Index \citep{DB} is also based on separation and compactness, although unlike the previous two measures, it is not given by the ratio of two values measuring these quantities.
Let \(\delta_j\) be the average distance of points in \(C_i\) to the center \(c_i\) of \(C_i\) (compactness) and let \(\Delta_{ij}\) be the distance between the centers \(c_i\) and \(c_j\) (separation) \citep{LiuEnh}:
\begin{align*} 
  \delta_i &= \frac{1}{n_i} \sum_{x \in C_i} d(x, c_i), \\
  \Delta_{ij} &= d(c_i, c_j).
\end{align*}
The similarity between two classes is given by \citep{LiuEnh}
\[
  s(C_i, C_j) = \frac{\delta_i + \delta_j}{\Delta_{ij}}.
\]
For each class, the maximum similarity is computed and the Davies-Bouldin index \(\\DB(X)\) is defined as the average of these maximum similarities:
\begin{definition}[Davies-Bouldin index]
\label{def:unnamed-chunk-46}
\begin{align*}
 &\DB(X) = \frac{1}{K} \sum_{i = 1}^{K} \max_{j, j\neq i} s(C_i, C_j) = \frac{1}{K} \sum_{i = 1}^{K} \max_{j, j\neq i} \frac{\delta_i + \delta_j}{\Delta_{ij}}, \\ 
 &\text{ where }\delta_i = \frac{1}{n_i} \sum_{x \in C_i} d(x, c_i) \text{ and } \Delta_{ij} = d(c_i, c_j) 
 \end{align*}
 \citep{LiuEnh}.
As the Davies-Bouldin index measures similarity between classes instead of distance or dissimilarity, smaller values indicate a better separation between classes.
In order to transform the values to \(]0,1]\) with \(1\) as best value, the DB index is modified as follows:
\({\displaystyle \DB(X)^* = \frac{1}{1 + \DB(X).}}\)
\end{definition}

\paragraph{Silhouette Index (Sil):}
The silhouette index \citep{Sil} is not based on a ratio of separation and compactness but on the differences of between- and within-cluster distances \citep{LiuEnh}.
First, a so called \emph{silhouette width} \(s(x)\) is calculated for each point \(x\):
Let \(a(x)\) be the average distance of a point \(x\) in class \(C_i\) to the other \(n_i - 1\) points in \(C_i\), let \(\delta(x, C_k)\) be the average distance to the points of another cluster \(C_k\) and let \(b(x)\) be the minimum of \(\delta(x, C_k)\) over all other classes \(k \neq i\), i.e., the minimum distance of \(x\) to another class; the ``second-best choice'' for \(x\) \citep{Sil}:
\begin{align*} 
  a(x) &= \frac{1}{n_i - 1} \sum_{x' \in C_i, x'\neq x} d(x, x') \text{ for } x \in C_i ,\\
  \delta(x, C_k) &= \frac{1}{n_k} \sum_{x' \in C_k} d(x, x') ,\\
  b(x) &= \min_{k = 1, ..., K, k\neq i}\delta(x, C_k) \text{ for } x \in C_i .
\end{align*}
 The silhouette width \(s(x)\) for each observation \(x\) is given by the following quotient \citep{Sil}:
\[
  s(x) = \frac{b(x) - a(x)}{\max\{ a(x), b(x) \}}.
\]
\(s(x)\) is between \(-1\) and \(1\) and indicates if \(x\) is assigned to the ``right'' cluster:
\(s(x)\) becomes \(1\) if \(a(x)\) is much smaller than \(b(x)\) which means that the average distance to the second-best choice (the class for which the minimum of \(\delta(x, C_k)\) is attained) is much higher than the average within-class distance \(a(x)\).
When \(s(x)\) is close to zero, this means that \(a(x)\) and \(b(x)\) have approximately the same value, i.e., \(x\) lies equally far from both its actual class and the second best choice.
The worst situation is a silhouette width close to \(-1\) which indicates that \(a(x)\) is much bigger than \(b(x)\), so \(x\) is much closer to the second-best choice than to its actual class \citep{Sil}.

The \(s(x)\) of all points can be plotted and used for graphical evaluations of clusterings \citep{Sil}.
In order to obtain a single value \(\Sil(X)\) that indicates the goodness of a given clustering (or given classes), one computes the mean silhouette width of each cluster and takes the mean of these values:
\begin{definition}[Silhouette index]
\label{def:unnamed-chunk-48}
\begin{align*}
 & \Sil(X) = \frac{1}{K} \sum_{i = 1}^{K} \frac{1}{n_i} \sum_{x \in C_i} \frac{b(x) - a(x)}{\max\{ a(x), b(x) \}} ,\\
 &\text{where } a(x) =\frac{1}{n_i - 1} \sum_{x' \in C_i, x'\neq x} d(x, x') \text{ and } b(x) = \min_{k = 1, ..., K, k\neq i} \left( \frac{1}{n_k} \sum_{x' \in C_k} d(x, x') \right) \text{ for } x \in C_i
\end{align*} \citep{LiuEnh}.
As \(\Sil(X) \in [-1, 1]\) and higher values indicate a better separation, the silhouette index is transformed to \([0, 1]\) as follows:
\({\displaystyle \Sil(X)^* = \frac{\Sil(X) + 1}{2}.}\)
\end{definition}

\paragraph{CVNN:}
The CVNN (clustering validation index based on nearest neighbors) \citep{LiuEnh} is a CVI that aims to overcome some limitations of existing CVIs.
As it was developed for clustering evaluation and it is based on notions of separation and compactness, the CVNN is presented in this section and not together with other measures that also use nearest neighbors in Section \ref{sec:MeasuresGraph}.
As mentioned above, most CVIs (including those presented in this section) cannot handle clusters of arbitrary shape \citep{LiuEnh}.
One reason for that is that many indices measure separation based on representatives of clusters, e.g the cluster center like the DB and CH index \citep{LiuEnh}.
The CVNN uses nearest neighbors to evaluate separation:
Let \(k\) be a number of nearest neighbors (e.g., \(k=10\)) and denote by \(q(x)\) the number of \(k\) nearest neighbors of \(x\) in class \(C_i\) that are not in \(C_i\).
Separation is defined as the maximum average proportion of nearest neighbors in other clusters.
The compactness within classes is given by the sum of average pairwise distance between points in the same class \citep{LiuEnh}:
\begin{align*} 
  \Sep_{\CVNN}(X) &= \max_{i = 1, ..., K} \frac{1}{n_i} \sum_{x \in C_i} \frac{q(x)}{k} ,\\
  \Comp_{\CVNN}(C_i) &= \frac{2}{n_i \cdot (n_i - 1)} \sum_{x, x' \in C_i} d(x,x') ,\\
  \Comp_{\CVNN}(X) &= \sum_{i = 1}^{K} Comp_{\CVNN}(C_i) = \sum_{i = 1}^{K} \frac{2}{n_i \cdot (n_i - 1)} \sum_{x, x' \in C_i} d(x,x') ,
\end{align*}
(the factor \(\frac{2}{n_i \cdot (n_i - 1)}\) is the inverse number of pairwise distances \(d(x,x')\) for \(x, x' \in C_i, x \neq x'\)).
The lower the value of \(\Sep_{\CVNN}\), the better the separation between classes.
Smaller values of \(\Comp_{\CVNN}\) indicate a better intra-class compactness.
\citet{LiuEnh} normalize both \(\Sep_{\CVNN}\) and \(\Comp_{\CVNN}\) to \([0, 1]\) and add them up in order to obtain a single value (i.e., \(\CVNN(X) = \Sep_{\CVNN, \norm} + \Comp_{\CVNN, \norm}\)).
The smaller the CVNN, the better.
As normalization factor, they use the maximum value of \(\Sep_{\CVNN}\) and \(\Comp_{\CVNN}\) among clustering results with different numbers \(K\) of clusters \citep{LiuEnh}.
While this makes sense when comparing clusterings for different numbers of clusters, this is not possible when the CVNN is used as a separability measure, as there are no partitions for different numbers of classes available.

The modified version of CVNN used in this paper is defined as follows:
With the above definition of \(\Comp_{\CVNN}\), this value depends highly on the scale of the distances in the data sets.
The modified compactness is given by the mean of \(\Comp_{\CVNN}(C_i)\) (instead of the sum) normalized by the mean pairwise distance in the data set:
\[
  \Comp_{\CVNN}(X)^* = \frac{\frac{1}{K} \sum_{i = 1}^{K} \frac{2}{n_i \cdot (n_i - 1)} \sum_{x, x' \in C_i} d(x,x'))}{\frac{2}{n \cdot (n-1)} \sum_{x, x' \in X} d(x,x')}.
\]
Now, the sum of \(\Comp_{\CVNN}(X)^*\) and \(\Sep_{\CVNN}(X)\) is transformed to \(]0, 1]\) with \(1\) as best value:
\begin{definition}[Modified CVNN index]
\label{def:unnamed-chunk-50}

\begin{align*}
&\CVNN(X)^* = \frac{1}{1 + \Comp_{\CVNN}(X)^* + \Sep_{\CVNN}(X)}, \\
&\text{where } \Comp_{\CVNN}(X)^* = \left(\frac{1}{K} \sum_{i = 1}^{K} \frac{2}{n_i \cdot (n_i - 1)} \sum_{x, x' \in C_i} d(x,x'))\right)/\left(\frac{2}{n \cdot (n-1)} \sum_{x, x' \in X} d(x,x')\right) \\
&\text{and } \Sep_{\CVNN}(X) = \max_{i = 1, ..., K} \frac{1}{n_i} \sum_{x \in C_i} \frac{q(x)}{k},
\end{align*}
 where \(q(x)\) denotes the number of \(k\) nearest neighbors of \(x\) that are not in the same class as \(x\).
\end{definition}

There are further attempts to develop CVIs that are able to deal with non-spherical clusters, for example the CVDD (cluster validity index based on density-involved distance) by \citet{Hu}.
Their notion of compactness uses path-based distances \citep{PBC} and is somewhat related to the idea of connectedness used for the separability measure proposed in Section \ref{sec:DCSI}.
The definition of separation in \citet{Hu} aims to be robust to outliers and to be able to cope with density-separated clusters as well as distance-separated cluster, whereas existing CVIs usually favor the latter.

\subsection{Distributional Approaches\label{sec:MeasuresDSI}}

\paragraph{DSI:}
The approach of \citet{GuanNN} to separability is different than the one of classical CVIs, as it is mainly based on the perspective of classification.
However, their \emph{distance-based separability index DSI} (DSI) can also be used for cluster validation \citep{GuanCVI}.
The DSI is based on the idea that the most difficult situation to separate is when two classes mix with each other, i.e., have the same distribution \citep{GuanNN}.
Consequently, separability can be defined in terms of the similarity of the distributions in different classes.
However, as the dimensions of these distributions can be very high, the idea of \citet{GuanNN} is to consider (one-dimensional) sets of pairwise distances.
Let \(\ICD(C_i)\) be the set of intra-class distances, i.e., the set of distances between any two points of \(C_i\), and let \(\BCD(C_i)\) be the set of between-class distances, i.e., the set of distances between any two points \(x, x'\) where \(x\in C_i, x'\notin C_i\) \citep{GuanNN}:
\begin{align*} 
  \ICD(C_i) &= \{ d(x, x'): x, x' \in C_i, x\neq x' \}, \\
  \BCD(C_i) &= \{ d(x, x'): x \in C_i, x' \notin C_i \} .
\end{align*}
Note that these ``sets'' are multisets, i.e., they can have duplicate elements (here distances) \citep{GuanSep}.
\citet{GuanNN} show that when \(n_i, n_j \rightarrow \infty\), if and only if two classes \(C_i\) and \(C_j\) have the same distribution, the distribution of the ICD and BCD sets is identical (in the case of two classes, so \(\BCD(C_i) = \{ d(x, x'): x \in C_i, x' \in C_j \}\)).
So instead of measuring the similarity of the original distributions, one examines the ICD and BCD sets.
\citet{GuanNN} apply the Kolmogorov-Smirnov test (KS) to compare the distributions of the ICD and BCD sets \(\ICD(C_i), \BCD(C_i)\) and measure their dissimilarity \(d(C_i)\).
The KS test is the maximum distance between two cumulative distribution functions (CDFs).
Let \(F_{\ICD_i}\) and \(F_{\BCD_i}\) be the CDFs of \(\ICD(C_i)\) and \(\BCD(C_i)\). Then \(d(C_i)\) is given by \citep{GuanNN}
\[ 
  d(C_i) = KS(\ICD(C_i), \BCD(C_i)) = \sup_x |F_{\ICD_i}(x) - F_{\BCD_i}(x)|.
\]
An alternative would be to use the Wasserstein distance \(W(\ICD(C_i), \BCD(C_i)) = \int |F_{\ICD_i}(x) - F_{\BCD_i}(x)| dx\) instead of the KS test, but \citet{GuanNN} find that the Wasserstein distance is less sensitive in measuring separability.
Higher values of \(d(C_i)\) (i.e., close to 1) indicate that class \(C_i\) is well separated from the others, as the distribution of the ICD and BCD set are very different.
The distance-based separability index (DSI) is defined as the mean of the \(d(C_i)\):
\begin{definition}[DSI]
\label{def:unnamed-chunk-51}
\begin{align*}
 &\DSI(X)  = \frac{\sum_{i = 1}^K d(C_i)}{K} , \\
&\text{where } d(C_i) = KS(\ICD(C_i), \BCD(C_i)) = \sup_x |F_{\ICD_i}(x) - F_{\BCD_i}(x)| \\
&\text{and } \ICD(C_i) = \{ d(x, x'): x, x' \in C_i, x\neq x' \},  \BCD(C_i) = \{ d(x, x'): x \in C_i, x' \notin C_i \}
\end{align*} \citep{GuanNN}.
\end{definition}
The DSI is between 0 and 1 and higher values indicate a higher separability.

There are many other ways to measure the similarity of distributions, e.g., divergence measures like the Jensen-Shannon divergence \citep{Lin}, however all approaches based on similarity of distributions only quantify separation but not connectedness.

\subsection{Graph- \& Neighborhood-Based Approaches\label{sec:MeasuresGraph}}
\label{sec:AppGraph}

This section presents measures from the categories \emph{neighborhood measures} and \emph{network measures} in \citet{Lorena}.
Neighborhood measures quantify the presence of points of different classes in local neighborhoods.
Network measures model the data as a graph and extract information from it.
Many neighborhood-based approaches can also be interpreted as graph-based, as some of these measures can also be extracted from (weighted) \(k\)-NN graphs or involve the construction of a particular graph or tree (like N1), so these two categories are combined in one section.
The first four measures (N1, N2, N3, LSC) are neighborhood measures.
The last two measures (Density and ClsCoef) are network measures.
They are both extracted from an \(\varepsilon\)-NN graph, i.e., a graph where two points \(x, x'\) are connected if and only if \(d(x, x')<\varepsilon\).
\citet{Lorena} use the Gower distance \citep{Gower} for both the neighborhood and the network measures, so in this section, \(d(x, x')\) denotes the Gower distance (however, all these measures can also be used with the Euclidean or any other distance instead).
The Gower distance is some kind of normalized Manhattan distances and takes values between 0 and 1 \citep{Gower, Lorena}.
To build the \(\varepsilon\)-NN graph, \(\varepsilon\) is set to \(0.15\) in \citet{Lorena}.
Then, the resulting graph is pruned: each edge between observations of different classes is removed \citep{Lorena}.
The pruned graph is used to extract measures of complexity or separability:
The more edges are removed, the lower is the separability.
The final graph is denoted by \(G = (V, E)\), where \(|V| = n\) and \(0 \leq |E| \leq \frac{n\cdot(n-1)}{2}\). \(v_i\) is the \(i\)-th vertex and an edge between \(v_i\) and \(v_j\) is denoted by \(e_{ij}\).

The complexity measures from \citet{Lorena} are all in \([0,1]\) with 1 indicating the highest possible complexity, i.e., lowest separability.
Here, each complexity measure \(C(X)\) is presented as \(1-C(X)\).
All definitions are taken from \citet{Lorena}.
Some of them can also be found in \citet{Ho2002}.

\paragraph{Fraction of Borderline Points (N1):}
To obtain this measure, one first builds a minimum spanning tree (MST)  from the data.
One then computes the percentage of observations that are connected to points from other classes (borderline points, here denoted by \(\operatorname{Bord}(X)\)).
Such points are either on the border or in regions with overlapping classes or noise that is surrounded by points from a different class.
So the higher the percentage of such points, the lower the separability.
Let \((x, x') \in \MST(X)\) denote that the points \(x, x'\) are connected by an edge in the MST build from the data \(X\) and let \(|\operatorname{Bord}(X)|\) be the cardinality of \(\operatorname{Bord}(X)\).
The separability measure \(\NOne(X)\) is given by the proportion of non-borderline points:
\begin{definition}[Fraction of borderline points (N1)]
\label{def:unnamed-chunk-52}
\[ \NOne(X) = 1 - \frac{1}{n} |\operatorname{Bord}(X)| ,\]

where \({\displaystyle x_i \in \operatorname{Bord}(X) \iff \exists x_j \in X: (x_i, x_j) \in \MST(X) \land y_i \neq y_j}\) \citep{Lorena}.
\end{definition}

\paragraph{Ratio of Intra/Extra Class Nearest Neighbor Distance (N2):}
For N2, one compares the sum of distances between each point \(x_i\) and its closest neighbor from the same class (\(\min_j\{ d(x_i, x_j)| y_i = y_j\}\)) and the sum of distances between each point and its closest neighbor from a different class (\(\min_j\{ d(x_i, x_j)| y_i \neq y_j\}\)):
\begin{definition}[Ratio of intra/extra class nearest neighbor distance (N2)]
\label{def:unnamed-chunk-53}
\[ \NTwo(X) = \frac{1}{1 + \operatorname{intra\_extra}(X)}, \]

where \({\displaystyle \operatorname{intra\_extra}(X) = \frac{\sum_{x_i \in X} \min_j\{ d(x_i, x_j)| y_i = y_j\}}{\sum_{x_i \in X} \min_j\{ d(x_i, x_j)| y_i \neq y_j\} }}\) \citep{Lorena}.
\end{definition}

\paragraph{Error Rate of the Nearest Neighbor Classifier (N3):}
N3 is computed from the error rate of a 1-nearest neighbor classifier using a leave-one-out estimate:
\begin{definition}[Error rate of the nearest neighbor classifier (N3)]
\label{def:unnamed-chunk-54}
\[  \NThree(X) = 1 - \frac{1}{n} |\operatorname{Err}_{NN}(X)| ,\]
where \({\displaystyle x_i \in \operatorname{Err}_{NN}(X) \iff NN(x_i) \neq y_i}\) and \(NN(x_i)\) is the predicted label from a 1-NN classifier \citep{Lorena}.
\end{definition}
 \(|\operatorname{Err}_{NN}(X)|\) denotes the cardinality of \(\operatorname{Err}_{NN}(X)\), the set of points in \(X\) that are misclassified using a 1-NN classifier.

\paragraph{Local Set Average Cardinality (LSC):}
For LSC, one considers the cardinality of so-called \emph{Local Sets} LS:
The LS of an observation \(x_i\) is defined as the set of points \(x_j\) that are closer to \(x_i\) than \(x_i\)'s closest neighbor from a different class.
The local set average cardinality is then given by
\begin{definition}[Local set average cardinality (LSC)]
\label{def:unnamed-chunk-55}
\[ LSC(X) = \frac{1}{n^2} \sum_{x \in X} |LS(x)|, \]
where \({\displaystyle LS(x_i) = \{ x_j| d(x_i, x_j) < \min_l\{ d(x_i, x_l)| y_i \neq y_l\} \}}\) \citep{Lorena}.
\end{definition}
In the ``least separable'' case, each observation \(x_i\) is closest to a point from a different class, so each local set has a cardinality of 1 (as it contains only \(x_i\)), resulting in a LSC of \(1/n\).
High values of LSC indicate that the classes are well separated from each other.
Note that the maximum possible value of LSC depends on the sizes of the classes.

\paragraph{Average density of the network (Density):}
This network measure is the number of edges in the final (i.e., pruned) graph divided by the maximum number of edges that can exist between \(n\) points (\(n\cdot(n-1)/2\)):
\begin{definition}[Average density of the network (Density)]
\label{def:unnamed-chunk-56}
\[  \Density(X) = \frac{2|E|}{n\cdot(n-1)} \quad  \text{\citep{Lorena}.}\]
\end{definition}
A dense graph (i.e., high values of \(|E|\)) indicates that there are dense regions within classes, so the separability is high \citep{Lorena}.

\paragraph{Clustering coefficient (ClsCoef):}
This network measure quantifies how much vertices of the same class form cliques:
For each vertex (i.e., observation) \(v_i\), one calculates the ratio of the number of edges between its neighbors and the maximum number of edges that could exist between them \citep{Lorena}.
\(N_i = \{ v_j: e_{ij} \in E\}\) denotes the neighborhood set of \(v_i\) and \(k_i\) is the size of \(N_i\), so there are \(k_i \cdot(k_i - 1)/2\) possible edges between the neighbors of \(v_i\).
\(|\{e_{jk}| v_j, v_k \in N_i \}|\) is the number of existing edges between neighbors of \(v_i\).
The clustering coefficient (ClsCoef) is the average proportion of existing edges:
\begin{definition}[Clustering coefficient (ClsCoef)]
\label{def:unnamed-chunk-57}
\[
 \ClsCoef(X) = \frac{1}{n} \sum_{i = 1}^n \frac{2 |\{e_{jk}| v_j, v_k \in N_i \}|}{k_i \cdot(k_i - 1)},
 \]
where \({\displaystyle N_i = \{ v_j: e_{ij} \in E\}}\) and \(k_i = |N_i|\) \citep{Lorena}.
\end{definition}

There are some other complexity or separability measures that can be found in literature.
The separability index (SI) by \citet{Thorn} is the same as N3 (both can also be extended to more neighbors than just one) \citep{Lorena}.
A measure called \emph{Hypothesis margin} (HM) \citep{Mth2} is similar to N2, as it compares distances to the nearest neighbor of the same class with distances to the nearest neighbor of a different class \citep{Lorena}.
\citet{Mth2} combine HM and Thornton's SI to a new hybrid measure that is able to differentiate between situations with a SI of \(100 \%\) (i.e., situations where no observation has a nearest neighbor from a different class).

The idea by \citet{Zighed} is somewhat similar to the network measures:
One first builds a graph that connects nearby observations, however they do not use an \(\varepsilon\)-NN or \(k\)-NN graph but a so-called ``Relative Neighborhood Graph'' (RNG) that contains a vertex between \(x_i\) and \(x_j\) if and only if the intersection of two hyperspheres centered on \(x_i\) and \(x_j\) with radius \(d(x_i,x_j)\) is empty \citep{Zighed}.
The next step is similar to the pruning-step in \citet{Lorena}: all edges that connect observations from different classes are removed.
Then, the relative weight of the removed edges (the ``cut edge weight statistic'') is computed.
\citet{Zighed} derive the distribution of this statistic under the null hypothesis \(H_0\) that the labels are assigned randomly and then calculate the p-value to evaluate the separability.
Similar to most other neighborhood- and graph-based measures, this approach doesn't quantify connectedness but only separation from a classification based view.
 
\if1\suppl
{

\pagebreak

\begin{center}
{\large\bf SUPPLEMENTARY MATERIAL}
\end{center}
 
\section{Definitions of existing separability measures}
\label{App:Def}
As it is desirable that all measures are in \([0, 1]\) (or \([0, 1[\) or \(]0, 1]\) etc.) with \(1\) as best value (highest separability), some measures are slightly modified which is indicated by an asterisk.
The notation is as follows: \(X=x_1, ..., x_n\) is a given data set with \(K\) classes \(C_1, ..., C_K\) of sizes \(n_1, ..., n_K\) with centers \(c_1, ..., c_K\) (i.e., the mean of each class).
\(c\) is the center of the whole data set.
\(d(x, x')\) denotes the Euclidean distance between \(x\) and \(x'\) (unless otherwise stated, see Section \ref{sec:MeasuresGraph}).
For some measures, a distance \(d(C_i, C_j)\) or a similarity \(s(C_i, C_j)\) between two classes \(C_i\) and \(C_j\) is defined.
\(\Sep(X)\) and \(\Comp(X)\) denote index specific definitions of separation and compactness.
For a point \(x_i\), \(y_i\) denotes the class label of \(x_i\).
For more details on the characteristics of the measures, see \citet{MA}.

\subsection{Internal Cluster Validity Indices\label{sec:MeasuresCVIs}}

\paragraph{Dunn Index:}
The Dunn Index \citep{Dunn} is the ratio of separation and compactness, which are defined as follows:
The distance between two classes \(C_j\) and \(C_j\) is the minimum distance between points of these classes.
The separation \(\Sep_{\Dunn}(X)\) of the whole data set \(X\) is given by the minimum distance between two classes \citep{Dunn}.
For a class $C_k$, the diameter \(\operatorname{diam}(C_k)\) is the maximum distance of points in this class.
The compactness \(\Comp_{\Dunn}(X)\) is given by the maximum diameter \citep{Dunn}:
\begin{align*} 
  d(C_i, C_j) & = \min_{x \in C_i, x' \in C_j} d(x,x') ,\\
  \Sep_{\Dunn}(X) &= \min_{i, j = 1, ..., K, i \neq j} d(C_i, C_j) ,\\
  \operatorname{diam}(C_k) &= \max_{x, x' \in C_k} d(x,x') ,\\
  \Comp_{\Dunn}(X) &= \max_{k = 1, ..., K} \operatorname{diam}(C_k).
\end{align*}
The Dunn index is the ratio of \(\Sep_{\Dunn}\) and \(\Comp_{\Dunn}\):
\begin{definition}[Dunn index]
\label{def:unnamed-chunk-42}
\[
\Dunn(X) = \frac{\Sep_{\Dunn}(X)}{\Comp_{\Dunn}(X)}= \frac{ \min_{i, j, i \neq j} (\min_{x \in C_i, x' \in C_j} d(x,x'))}{\max_k (\max_{x, x' \in C_k} d(x,x'))}\] \citep{Dunn}.
As there is no upper limit, the Dunn index is slightly modified to be in \([0,1[\):
\[
\Dunn(X)^* = \frac{\Dunn(X)}{1 + \Dunn(X)}.
\]
\end{definition}

\paragraph{Calinski-Harabasz Index (CH):}
The Calinski-Harabasz Index (CH) \citep{CH} also takes the form \(\Sep_{\CH}/\Comp_{\CH}\) \citep{LiuEnh}.
Separation is measured in terms of the weighted sum of squared distances of the class centers to the center \(c\) of the whole data set.
Compactness is based on the within-group variance \citep{LiuEnh}:
\begin{align*} 
  \Sep_{\CH}(X) &= \frac{1}{K - 1} \sum_{i = 1}^{K} n_i d(c_i, c)^2 ,\\
  \Comp_{\CH}(X) &= \frac{1}{n - K} \sum_{i = 1}^{K} \sum_{x \in C_i} d(x, c_i)^2.
\end{align*}
The CH index is defined as
\begin{definition}[Calinski-Harabasz index]
\label{def:unnamed-chunk-44}
\[
\CH(X) = \frac{\Sep_{\CH}(X)}{\Comp_{\CH}(X)}= \frac{\sum_{i = 1}^{K} n_i d(c_i, c)^2 / (K - 1)}{\sum_{i = 1}^{K} \sum_{x \in C_i} d(x, c_i)^2 / (n - K)} = \frac{n - K}{K - 1} \frac{\sum_{i = 1}^{K} n_i d(c_i, c)^2)}{\sum_{i = 1}^{K} \sum_{x \in C_i} d(x, c_i)^2}
\] \citep{ LiuEnh}.
CH can take arbitrary high values.
The modified version of the CH index is given by
\[
 \CH(X)^* = \frac{\CH(X)^{**}}{1 + \CH(X)^{**}}
\text{ where }\CH(X)^{**} = \frac{K-1}{n-K}\CH(X).
\]
\end{definition}
As this index is used as a CVI and the term \(\sum_{i = 1}^{K} \sum_{x \in C_i} d(x, c_i)^2\) (typically) becomes smaller as the number \(K\) of clusters increases, it is corrected by multiplying with \((n-K)/(K-1)\), which decreases as \(K\) increases.
However, when used as a separability measure, no correction for the number of classes is needed.

\paragraph{Davies-Bouldin Index (DB):}
The Davies-Bouldin Index \citep{DB} is also based on separation and compactness, although unlike the previous two measures, it is not given by the ratio of two values measuring these quantities.
Let \(\delta_j\) be the average distance of points in \(C_i\) to the center \(c_i\) of \(C_i\) (compactness) and let \(\Delta_{ij}\) be the distance between the centers \(c_i\) and \(c_j\) (separation) \citep{LiuEnh}:
\begin{align*} 
  \delta_i &= \frac{1}{n_i} \sum_{x \in C_i} d(x, c_i), \\
  \Delta_{ij} &= d(c_i, c_j).
\end{align*}
The similarity between two classes is given by \citep{LiuEnh}
\[
  s(C_i, C_j) = \frac{\delta_i + \delta_j}{\Delta_{ij}}.
\]
For each class, the maximum similarity is computed and the Davies-Bouldin index \(\\DB(X)\) is defined as the average of these maximum similarities:
\begin{definition}[Davies-Bouldin index]
\label{def:unnamed-chunk-46}
\begin{align*}
 &\DB(X) = \frac{1}{K} \sum_{i = 1}^{K} \max_{j, j\neq i} s(C_i, C_j) = \frac{1}{K} \sum_{i = 1}^{K} \max_{j, j\neq i} \frac{\delta_i + \delta_j}{\Delta_{ij}}, \\ 
 &\text{ where }\delta_i = \frac{1}{n_i} \sum_{x \in C_i} d(x, c_i) \text{ and } \Delta_{ij} = d(c_i, c_j) 
 \end{align*}
 \citep{LiuEnh}.
As the Davies-Bouldin index measures similarity between classes instead of distance or dissimilarity, smaller values indicate a better separation between classes.
In order to transform the values to \(]0,1]\) with \(1\) as best value, the DB index is modified as follows:
\({\displaystyle \DB(X)^* = \frac{1}{1 + \DB(X).}}\)
\end{definition}

\paragraph{Silhouette Index (Sil):}
The silhouette index \citep{Sil} is not based on a ratio of separation and compactness but on the differences of between- and within-cluster distances \citep{LiuEnh}.
First, a so called \emph{silhouette width} \(s(x)\) is calculated for each point \(x\):
Let \(a(x)\) be the average distance of a point \(x\) in class \(C_i\) to the other \(n_i - 1\) points in \(C_i\), let \(\delta(x, C_k)\) be the average distance to the points of another cluster \(C_k\) and let \(b(x)\) be the minimum of \(\delta(x, C_k)\) over all other classes \(k \neq i\), i.e., the minimum distance of \(x\) to another class; the ``second-best choice'' for \(x\) \citep{Sil}:
\begin{align*} 
  a(x) &= \frac{1}{n_i - 1} \sum_{x' \in C_i, x'\neq x} d(x, x') \text{ for } x \in C_i ,\\
  \delta(x, C_k) &= \frac{1}{n_k} \sum_{x' \in C_k} d(x, x') ,\\
  b(x) &= \min_{k = 1, ..., K, k\neq i}\delta(x, C_k) \text{ for } x \in C_i .
\end{align*}
 The silhouette width \(s(x)\) for each observation \(x\) is given by the following quotient \citep{Sil}:
\[
  s(x) = \frac{b(x) - a(x)}{\max\{ a(x), b(x) \}}.
\]
\(s(x)\) is between \(-1\) and \(1\) and indicates if \(x\) is assigned to the ``right'' cluster:
\(s(x)\) becomes \(1\) if \(a(x)\) is much smaller than \(b(x)\) which means that the average distance to the second-best choice (the class for which the minimum of \(\delta(x, C_k)\) is attained) is much higher than the average within-class distance \(a(x)\).
When \(s(x)\) is close to zero, this means that \(a(x)\) and \(b(x)\) have approximately the same value, i.e., \(x\) lies equally far from both its actual class and the second best choice.
The worst situation is a silhouette width close to \(-1\) which indicates that \(a(x)\) is much bigger than \(b(x)\), so \(x\) is much closer to the second-best choice than to its actual class \citep{Sil}.

The \(s(x)\) of all points can be plotted and used for graphical evaluations of clusterings \citep{Sil}.
In order to obtain a single value \(\Sil(X)\) that indicates the goodness of a given clustering (or given classes), one computes the mean silhouette width of each cluster and takes the mean of these values:
\begin{definition}[Silhouette index]
\label{def:unnamed-chunk-48}
\begin{align*}
 & \Sil(X) = \frac{1}{K} \sum_{i = 1}^{K} \frac{1}{n_i} \sum_{x \in C_i} \frac{b(x) - a(x)}{\max\{ a(x), b(x) \}} ,\\
 &\text{where } a(x) =\frac{1}{n_i - 1} \sum_{x' \in C_i, x'\neq x} d(x, x') \text{ and } b(x) = \min_{k = 1, ..., K, k\neq i} \left( \frac{1}{n_k} \sum_{x' \in C_k} d(x, x') \right) \text{ for } x \in C_i
\end{align*} \citep{LiuEnh}.
As \(\Sil(X) \in [-1, 1]\) and higher values indicate a better separation, the silhouette index is transformed to \([0, 1]\) as follows:
\({\displaystyle \Sil(X)^* = \frac{\Sil(X) + 1}{2}.}\)
\end{definition}

\paragraph{CVNN:}
The CVNN (clustering validation index based on nearest neighbors) \citep{LiuEnh} is a CVI that aims to overcome some limitations of existing CVIs.
As it was developed for clustering evaluation and it is based on notions of separation and compactness, the CVNN is presented in this section and not together with other measures that also use nearest neighbors in Section \ref{sec:MeasuresGraph}.
As mentioned above, most CVIs (including those presented in this section) cannot handle clusters of arbitrary shape \citep{LiuEnh}.
One reason for that is that many indices measure separation based on representatives of clusters, e.g the cluster center like the DB and CH index \citep{LiuEnh}.
The CVNN uses nearest neighbors to evaluate separation:
Let \(k\) be a number of nearest neighbors (e.g., \(k=10\)) and denote by \(q(x)\) the number of \(k\) nearest neighbors of \(x\) in class \(C_i\) that are not in \(C_i\).
Separation is defined as the maximum average proportion of nearest neighbors in other clusters.
The compactness within classes is given by the sum of average pairwise distance between points in the same class \citep{LiuEnh}:
\begin{align*} 
  \Sep_{\CVNN}(X) &= \max_{i = 1, ..., K} \frac{1}{n_i} \sum_{x \in C_i} \frac{q(x)}{k} ,\\
  \Comp_{\CVNN}(C_i) &= \frac{2}{n_i \cdot (n_i - 1)} \sum_{x, x' \in C_i} d(x,x') ,\\
  \Comp_{\CVNN}(X) &= \sum_{i = 1}^{K} Comp_{\CVNN}(C_i) = \sum_{i = 1}^{K} \frac{2}{n_i \cdot (n_i - 1)} \sum_{x, x' \in C_i} d(x,x') ,
\end{align*}
(the factor \(\frac{2}{n_i \cdot (n_i - 1)}\) is the inverse number of pairwise distances \(d(x,x')\) for \(x, x' \in C_i, x \neq x'\)).
The lower the value of \(\Sep_{\CVNN}\), the better the separation between classes.
Smaller values of \(\Comp_{\CVNN}\) indicate a better intra-class compactness.
\citet{LiuEnh} normalize both \(\Sep_{\CVNN}\) and \(\Comp_{\CVNN}\) to \([0, 1]\) and add them up in order to obtain a single value (i.e., \(\CVNN(X) = \Sep_{\CVNN, \norm} + \Comp_{\CVNN, \norm}\)).
The smaller the CVNN, the better.
As normalization factor, they use the maximum value of \(\Sep_{\CVNN}\) and \(\Comp_{\CVNN}\) among clustering results with different numbers \(K\) of clusters \citep{LiuEnh}.
While this makes sense when comparing clusterings for different numbers of clusters, this is not possible when the CVNN is used as a separability measure, as there are no partitions for different numbers of classes available.

The modified version of CVNN used in this paper is defined as follows:
With the above definition of \(\Comp_{\CVNN}\), this value depends highly on the scale of the distances in the data sets.
The modified compactness is given by the mean of \(\Comp_{\CVNN}(C_i)\) (instead of the sum) normalized by the mean pairwise distance in the data set:
\[
  \Comp_{\CVNN}(X)^* = \frac{\frac{1}{K} \sum_{i = 1}^{K} \frac{2}{n_i \cdot (n_i - 1)} \sum_{x, x' \in C_i} d(x,x'))}{\frac{2}{n \cdot (n-1)} \sum_{x, x' \in X} d(x,x')}.
\]
Now, the sum of \(\Comp_{\CVNN}(X)^*\) and \(\Sep_{\CVNN}(X)\) is transformed to \(]0, 1]\) with \(1\) as best value:
\begin{definition}[Modified CVNN index]
\label{def:unnamed-chunk-50}

\begin{align*}
&\CVNN(X)^* = \frac{1}{1 + \Comp_{\CVNN}(X)^* + \Sep_{\CVNN}(X)}, \\
&\text{where } \Comp_{\CVNN}(X)^* = \left(\frac{1}{K} \sum_{i = 1}^{K} \frac{2}{n_i \cdot (n_i - 1)} \sum_{x, x' \in C_i} d(x,x'))\right)/\left(\frac{2}{n \cdot (n-1)} \sum_{x, x' \in X} d(x,x')\right) \\
&\text{and } \Sep_{\CVNN}(X) = \max_{i = 1, ..., K} \frac{1}{n_i} \sum_{x \in C_i} \frac{q(x)}{k},
\end{align*}
 where \(q(x)\) denotes the number of \(k\) nearest neighbors of \(x\) that are not in the same class as \(x\).
\end{definition}

There are further attempts to develop CVIs that are able to deal with non-spherical clusters, for example the CVDD (cluster validity index based on density-involved distance) by \citet{Hu}.
Their notion of compactness uses path-based distances \citep{PBC} and is somewhat related to the idea of connectedness used for the separability measure proposed in Section \ref{sec:DCSI}.
The definition of separation in \citet{Hu} aims to be robust to outliers and to be able to cope with density-separated clusters as well as distance-separated cluster, whereas existing CVIs usually favor the latter.

\subsection{Distributional Approaches\label{sec:MeasuresDSI}}

\paragraph{DSI:}
The approach of \citet{GuanNN} to separability is different than the one of classical CVIs, as it is mainly based on the perspective of classification.
However, their \emph{distance-based separability index DSI} (DSI) can also be used for cluster validation \citep{GuanCVI}.
The DSI is based on the idea that the most difficult situation to separate is when two classes mix with each other, i.e., have the same distribution \citep{GuanNN}.
Consequently, separability can be defined in terms of the similarity of the distributions in different classes.
However, as the dimensions of these distributions can be very high, the idea of \citet{GuanNN} is to consider (one-dimensional) sets of pairwise distances.
Let \(\ICD(C_i)\) be the set of intra-class distances, i.e., the set of distances between any two points of \(C_i\), and let \(\BCD(C_i)\) be the set of between-class distances, i.e., the set of distances between any two points \(x, x'\) where \(x\in C_i, x'\notin C_i\) \citep{GuanNN}:
\begin{align*} 
  \ICD(C_i) &= \{ d(x, x'): x, x' \in C_i, x\neq x' \}, \\
  \BCD(C_i) &= \{ d(x, x'): x \in C_i, x' \notin C_i \} .
\end{align*}
Note that these ``sets'' are multisets, i.e., they can have duplicate elements (here distances) \citep{GuanSep}.
\citet{GuanNN} show that when \(n_i, n_j \rightarrow \infty\), if and only if two classes \(C_i\) and \(C_j\) have the same distribution, the distribution of the ICD and BCD sets is identical (in the case of two classes, so \(\BCD(C_i) = \{ d(x, x'): x \in C_i, x' \in C_j \}\)).
So instead of measuring the similarity of the original distributions, one examines the ICD and BCD sets.
\citet{GuanNN} apply the Kolmogorov-Smirnov test (KS) to compare the distributions of the ICD and BCD sets \(\ICD(C_i), \BCD(C_i)\) and measure their dissimilarity \(d(C_i)\).
The KS test is the maximum distance between two cumulative distribution functions (CDFs).
Let \(F_{\ICD_i}\) and \(F_{\BCD_i}\) be the CDFs of \(\ICD(C_i)\) and \(\BCD(C_i)\). Then \(d(C_i)\) is given by \citep{GuanNN}
\[ 
  d(C_i) = KS(\ICD(C_i), \BCD(C_i)) = \sup_x |F_{\ICD_i}(x) - F_{\BCD_i}(x)|.
\]
An alternative would be to use the Wasserstein distance \(W(\ICD(C_i), \BCD(C_i)) = \int |F_{\ICD_i}(x) - F_{\BCD_i}(x)| dx\) instead of the KS test, but \citet{GuanNN} find that the Wasserstein distance is less sensitive in measuring separability.
Higher values of \(d(C_i)\) (i.e., close to 1) indicate that class \(C_i\) is well separated from the others, as the distribution of the ICD and BCD set are very different.
The distance-based separability index (DSI) is defined as the mean of the \(d(C_i)\):
\begin{definition}[DSI]
\label{def:unnamed-chunk-51}
\begin{align*}
 &\DSI(X)  = \frac{\sum_{i = 1}^K d(C_i)}{K} , \\
&\text{where } d(C_i) = KS(\ICD(C_i), \BCD(C_i)) = \sup_x |F_{\ICD_i}(x) - F_{\BCD_i}(x)| \\
&\text{and } \ICD(C_i) = \{ d(x, x'): x, x' \in C_i, x\neq x' \},  \BCD(C_i) = \{ d(x, x'): x \in C_i, x' \notin C_i \}
\end{align*} \citep{GuanNN}.
\end{definition}
The DSI is between 0 and 1 and higher values indicate a higher separability.

There are many other ways to measure the similarity of distributions, e.g., divergence measures like the Jensen-Shannon divergence \citep{Lin}, however all approaches based on similarity of distributions only quantify separation but not connectedness.

\subsection{Graph- \& Neighborhood-Based Approaches\label{sec:MeasuresGraph}}
\label{sec:AppGraph}

This section presents measures from the categories \emph{neighborhood measures} and \emph{network measures} in \citet{Lorena}.
Neighborhood measures quantify the presence of points of different classes in local neighborhoods.
Network measures model the data as a graph and extract information from it.
Many neighborhood-based approaches can also be interpreted as graph-based, as some of these measures can also be extracted from (weighted) \(k\)-NN graphs or involve the construction of a particular graph or tree (like N1), so these two categories are combined in one section.
The first four measures (N1, N2, N3, LSC) are neighborhood measures.
The last two measures (Density and ClsCoef) are network measures.
They are both extracted from an \(\varepsilon\)-NN graph, i.e., a graph where two points \(x, x'\) are connected if and only if \(d(x, x')<\varepsilon\).
\citet{Lorena} use the Gower distance \citep{Gower} for both the neighborhood and the network measures, so in this section, \(d(x, x')\) denotes the Gower distance (however, all these measures can also be used with the Euclidean or any other distance instead).
The Gower distance is some kind of normalized Manhattan distances and takes values between 0 and 1 \citep{Gower, Lorena}.
To build the \(\varepsilon\)-NN graph, \(\varepsilon\) is set to \(0.15\) in \citet{Lorena}.
Then, the resulting graph is pruned: each edge between observations of different classes is removed \citep{Lorena}.
The pruned graph is used to extract measures of complexity or separability:
The more edges are removed, the lower is the separability.
The final graph is denoted by \(G = (V, E)\), where \(|V| = n\) and \(0 \leq |E| \leq \frac{n\cdot(n-1)}{2}\). \(v_i\) is the \(i\)-th vertex and an edge between \(v_i\) and \(v_j\) is denoted by \(e_{ij}\).

The complexity measures from \citet{Lorena} are all in \([0,1]\) with 1 indicating the highest possible complexity, i.e., lowest separability.
Here, each complexity measure \(C(X)\) is presented as \(1-C(X)\).
All definitions are taken from \citet{Lorena}.
Some of them can also be found in \citet{Ho2002}.

\paragraph{Fraction of Borderline Points (N1):}
To obtain this measure, one first builds a minimum spanning tree (MST)  from the data.
One then computes the percentage of observations that are connected to points from other classes (borderline points, here denoted by \(\operatorname{Bord}(X)\)).
Such points are either on the border or in regions with overlapping classes or noise that is surrounded by points from a different class.
So the higher the percentage of such points, the lower the separability.
Let \((x, x') \in \MST(X)\) denote that the points \(x, x'\) are connected by an edge in the MST build from the data \(X\) and let \(|\operatorname{Bord}(X)|\) be the cardinality of \(\operatorname{Bord}(X)\).
The separability measure \(\NOne(X)\) is given by the proportion of non-borderline points:
\begin{definition}[Fraction of borderline points (N1)]
\label{def:unnamed-chunk-52}
\[ \NOne(X) = 1 - \frac{1}{n} |\operatorname{Bord}(X)| ,\]

where \({\displaystyle x_i \in \operatorname{Bord}(X) \iff \exists x_j \in X: (x_i, x_j) \in \MST(X) \land y_i \neq y_j}\) \citep{Lorena}.
\end{definition}

\paragraph{Ratio of Intra/Extra Class Nearest Neighbor Distance (N2):}
For N2, one compares the sum of distances between each point \(x_i\) and its closest neighbor from the same class (\(\min_j\{ d(x_i, x_j)| y_i = y_j\}\)) and the sum of distances between each point and its closest neighbor from a different class (\(\min_j\{ d(x_i, x_j)| y_i \neq y_j\}\)):
\begin{definition}[Ratio of intra/extra class nearest neighbor distance (N2)]
\label{def:unnamed-chunk-53}
\[ \NTwo(X) = \frac{1}{1 + \operatorname{intra\_extra}(X)}, \]

where \({\displaystyle \operatorname{intra\_extra}(X) = \frac{\sum_{x_i \in X} \min_j\{ d(x_i, x_j)| y_i = y_j\}}{\sum_{x_i \in X} \min_j\{ d(x_i, x_j)| y_i \neq y_j\} }}\) \citep{Lorena}.
\end{definition}

\paragraph{Error Rate of the Nearest Neighbor Classifier (N3):}
N3 is computed from the error rate of a 1-nearest neighbor classifier using a leave-one-out estimate:
\begin{definition}[Error rate of the nearest neighbor classifier (N3)]
\label{def:unnamed-chunk-54}
\[  \NThree(X) = 1 - \frac{1}{n} |\operatorname{Err}_{NN}(X)| ,\]
where \({\displaystyle x_i \in \operatorname{Err}_{NN}(X) \iff NN(x_i) \neq y_i}\) and \(NN(x_i)\) is the predicted label from a 1-NN classifier \citep{Lorena}.
\end{definition}
 \(|\operatorname{Err}_{NN}(X)|\) denotes the cardinality of \(\operatorname{Err}_{NN}(X)\), the set of points in \(X\) that are misclassified using a 1-NN classifier.

\paragraph{Local Set Average Cardinality (LSC):}
For LSC, one considers the cardinality of so-called \emph{Local Sets} LS:
The LS of an observation \(x_i\) is defined as the set of points \(x_j\) that are closer to \(x_i\) than \(x_i\)'s closest neighbor from a different class.
The local set average cardinality is then given by
\begin{definition}[Local set average cardinality (LSC)]
\label{def:unnamed-chunk-55}
\[ LSC(X) = \frac{1}{n^2} \sum_{x \in X} |LS(x)|, \]
where \({\displaystyle LS(x_i) = \{ x_j| d(x_i, x_j) < \min_l\{ d(x_i, x_l)| y_i \neq y_l\} \}}\) \citep{Lorena}.
\end{definition}
In the ``least separable'' case, each observation \(x_i\) is closest to a point from a different class, so each local set has a cardinality of 1 (as it contains only \(x_i\)), resulting in a LSC of \(1/n\).
High values of LSC indicate that the classes are well separated from each other.
Note that the maximum possible value of LSC depends on the sizes of the classes.

\paragraph{Average density of the network (Density):}
This network measure is the number of edges in the final (i.e., pruned) graph divided by the maximum number of edges that can exist between \(n\) points (\(n\cdot(n-1)/2\)):
\begin{definition}[Average density of the network (Density)]
\label{def:unnamed-chunk-56}
\[  \Density(X) = \frac{2|E|}{n\cdot(n-1)} \quad  \text{\citep{Lorena}.}\]
\end{definition}
A dense graph (i.e., high values of \(|E|\)) indicates that there are dense regions within classes, so the separability is high \citep{Lorena}.

\paragraph{Clustering coefficient (ClsCoef):}
This network measure quantifies how much vertices of the same class form cliques:
For each vertex (i.e., observation) \(v_i\), one calculates the ratio of the number of edges between its neighbors and the maximum number of edges that could exist between them \citep{Lorena}.
\(N_i = \{ v_j: e_{ij} \in E\}\) denotes the neighborhood set of \(v_i\) and \(k_i\) is the size of \(N_i\), so there are \(k_i \cdot(k_i - 1)/2\) possible edges between the neighbors of \(v_i\).
\(|\{e_{jk}| v_j, v_k \in N_i \}|\) is the number of existing edges between neighbors of \(v_i\).
The clustering coefficient (ClsCoef) is the average proportion of existing edges:
\begin{definition}[Clustering coefficient (ClsCoef)]
\label{def:unnamed-chunk-57}
\[
 \ClsCoef(X) = \frac{1}{n} \sum_{i = 1}^n \frac{2 |\{e_{jk}| v_j, v_k \in N_i \}|}{k_i \cdot(k_i - 1)},
 \]
where \({\displaystyle N_i = \{ v_j: e_{ij} \in E\}}\) and \(k_i = |N_i|\) \citep{Lorena}.
\end{definition}

There are some other complexity or separability measures that can be found in literature.
The separability index (SI) by \citet{Thorn} is the same as N3 (both can also be extended to more neighbors than just one) \citep{Lorena}.
A measure called \emph{Hypothesis margin} (HM) \citep{Mth2} is similar to N2, as it compares distances to the nearest neighbor of the same class with distances to the nearest neighbor of a different class \citep{Lorena}.
\citet{Mth2} combine HM and Thornton's SI to a new hybrid measure that is able to differentiate between situations with a SI of \(100 \%\) (i.e., situations where no observation has a nearest neighbor from a different class).

The idea by \citet{Zighed} is somewhat similar to the network measures:
One first builds a graph that connects nearby observations, however they do not use an \(\varepsilon\)-NN or \(k\)-NN graph but a so-called ``Relative Neighborhood Graph'' (RNG) that contains a vertex between \(x_i\) and \(x_j\) if and only if the intersection of two hyperspheres centered on \(x_i\) and \(x_j\) with radius \(d(x_i,x_j)\) is empty \citep{Zighed}.
The next step is similar to the pruning-step in \citet{Lorena}: all edges that connect observations from different classes are removed.
Then, the relative weight of the removed edges (the ``cut edge weight statistic'') is computed.
\citet{Zighed} derive the distribution of this statistic under the null hypothesis \(H_0\) that the labels are assigned randomly and then calculate the p-value to evaluate the separability.
Similar to most other neighborhood- and graph-based measures, this approach doesn't quantify connectedness but only separation from a classification based view.
 }\fi

\end{document}